\documentclass[a4paper, 12pt]{article} 
\usepackage{apacite}
\usepackage{authblk}
\usepackage{amsmath, amsfonts, amssymb}
\usepackage[a4paper, margin=2.5 cm]{geometry}
\usepackage{natbib}
\usepackage{graphicx} 
\usepackage{longtable}
\usepackage{array}
\usepackage{booktabs} 
\usepackage[colorlinks=true, linkcolor=blue, citecolor=blue]{hyperref}
\usepackage{tabularx}
\usepackage{setspace}
\usepackage{subcaption}
\usepackage{threeparttable}
\usepackage{titling}
\usepackage{xcolor}
\newcommand{\subtitle}[1]{%
  \posttitle{%
    \par\end{center}
    \begin{center}\large#1\end{center}
    \vskip0.2em}%
}

\title{\textbf{Measuring Poverty and Inequality with Reduced Data}}
\subtitle{\large\textbf{A Machine Learning Approach Using Nigerian Household Data}}

\author{Vanesa Jordá\thanks{Department of Economics, Cantabria University. Email: vanesa.jorda@unican.es} \and Miguel Niño-Zarazúa\thanks{Department of Economics, SOAS University of London. Email: mn39@soas.ac.uk}\thanks{United Nations University World Institute for Development Economics Research (UNU-WIDER)}}

\date{}

\begin{document}

\maketitle

\setlength{\parskip}{10pt}

\begin{abstract}
Reliable measurement of income and consumption is essential for monitoring poverty and inequality in low- and middle-income countries, yet full household surveys are costly and difficult to implement regularly. This paper examines whether reduced survey instruments can preserve key distributional information. We apply Random Forest Recursive Feature Elimination (RF-RFE) to the 2018/19 Nigeria General Household Survey-Panel to identify the income sources, consumption categories and household characteristics that best classify individuals within the welfare distribution. The analysis focuses on three outcomes: poverty status, location in the quintile distribution and position relative to the Gini-based inequality line. The survey's post-planting and post-harvest periods allow us to assess performance under different seasonal contexts. Results show that RF-RFE achieves strong classification accuracy with few predictors. For consumption, poverty status and inequality-line position are accurately predicted using a small set of expenditure categories, while quintile classification reaches about 80 percent accuracy for seasonal consumption and 60--65 percent for annual consumption predicted from a single seasonal visit. For income, poverty status reaches around 90 percent accuracy with five predictors, and inequality-line position is largely captured by labour earnings. The findings suggest that machine-learning methods can help improve survey design and reduce data requirements while retaining much of the distributional information needed to measure and monitor poverty and inequality.
\end{abstract}



\section{Introduction}
\label{sec:Introduction}

Accurate measurement of income and consumption is central to poverty and inequality analysis. Income captures the flow of resources available to households through labour, agriculture, enterprises, transfers and other sources, while consumption expenditure reflects how resources are used to meet needs and sustain living standards. Together, these welfare aggregates allow researchers and policymakers to identify deprivation, monitor distributional change, and assess the extent to which economic resources are unequally distributed across individuals, households and communities.

Income and consumption serve distinct analytical purposes. Income data are generally more directly connected to inequality analysis because they capture command over resources at a given point in time and are closely aligned with internationally agreed statistical definitions, such as those established by the Canberra Group \citep{canberragroup2011}. Income is also closely linked to the distribution of earnings, capital income and transfers \citep{atkinson1970measurement,piketty2014capital}, and is therefore central to the study of inequality \citep{cowell2011measuring,bourguignon2015global}. By contrast, consumption expenditure is often preferred for poverty analysis, particularly in low-income settings, because it is typically less volatile than income and may better approximate longer-term living standards \citep{deaton2002guidelines,ravallion1998poverty}. Consumption may smooth temporary income fluctuations and can therefore provide a more reliable indicator of material deprivation, especially at the lower end of the welfare distribution.

In sub-Saharan Africa, household surveys have traditionally prioritised the collection of detailed consumption expenditure data rather than comprehensive income data. Large-scale survey programmes such as the Living Standards Measurement Study (LSMS) have adopted consumption aggregates as the primary welfare indicator \citep{beegle2012methods,deaton1997analysis}. This reflects both conceptual and practical considerations. In economies characterised by informality, self-employment and subsistence agriculture, income is difficult to measure accurately because it is irregular, seasonal and often derived from multiple activities. Detailed expenditure modules are also costly and time-consuming, but they tend to provide more stable and reliable welfare estimates. As a result, poverty measurement across the region is overwhelmingly consumption-based, while income data, where available, are often used in complementary analyses of earnings, livelihoods and inequality.

However, collecting income and consumption data through nationally representative household surveys requires substantial financial and human resources. Such surveys require trained enumerators, complex questionnaire modules, repeated visits in settings where seasonality matters, data collection technology, field supervision and statistical expertise.\footnote{The cost of implementing LSMS-type surveys in sub-Saharan Africa is significant. While costs vary by sample size, geography and survey design, the World Bank's Independent Evaluation Group estimates the average cost of a standard national household survey at approximately US\$1.7 million \citep{worldbank2023poverty}. On a unit basis, costs in African survey contexts can range from US\$100 to US\$500 per household, reflecting the high marginal costs of reaching remote rural populations and the increasing use of Computer-Assisted Personal Interviewing technologies \citep{kilic2017costing}. The global requirement for monitoring the Sustainable Development Goals through household surveys was projected at nearly US\$945 million between 2016 and 2030 \citep{kilic2017costing,grosh1998data}.} These costs limit the frequency with which full welfare surveys can be implemented, particularly in low- and middle-income countries with constrained statistical capacity. They also create a practical need for shorter instruments that can retain key distributional information while reducing respondent burden and fieldwork costs.

A growing body of literature has developed alternative strategies to approximate income or consumption using fewer survey questions. These include asset-based wealth indices constructed through principal component analysis \citep{FilmerPritchett2001,montgomery2000measuring,mckenzie2005measuring,vyas2006constructing}, supervised proxy means tests and poverty scorecards calibrated to predict poverty status \citep{GroshBaker1995,Schreiner2010}, shortened consumption modules based on selected expenditure categories \citep{BeegleEtAl2012,GibsonKim2007}, single or bracketed income questions designed to reduce reporting error \citep{Hurd2004,Kennickell1998}, and subjective welfare measures based on self-assessed economic status \citep{PradhanRavallion2000,RavallionLokshin2002}. These approaches reduce survey burden, but they are usually designed for specific purposes, such as poverty targeting, broad wealth ranking or rapid monitoring. They are not designed to preserve several distributional classifications simultaneously.

This paper contributes to the literature on welfare measurement by proposing a machine-learning approach to identify a parsimonious set of income sources and consumption categories that can classify individuals within the welfare distribution. Specifically, we implement Random Forest Recursive Feature Elimination (RF-RFE), a supervised feature-selection method that ranks candidate variables according to their predictive contribution and evaluates model performance across progressively smaller sets of predictors. Unlike standard proxy means-testing approaches, which are typically designed to identify households below a poverty threshold, our approach evaluates whether reduced survey instruments can preserve three complementary distributional outcomes: poverty status, quintile classification and position relative to the Gini-based inequality line.

The empirical analysis uses the fourth wave of the Nigeria General Household Survey-Panel, collected in 2018/19. Nigeria provides a relevant setting for this exercise because household welfare is shaped by agricultural seasonality, income volatility, regional heterogeneity and the coexistence of formal, informal and subsistence economic activities. The survey's post-planting and post-harvest visits allow us to examine whether reduced instruments perform differently across seasonal information sets. We construct benchmark income and consumption aggregates from the full survey modules, derive the corresponding poverty, quintile and inequality-line classifications, and then assess how accurately RF-RFE models reproduce these classifications using a much smaller set of predictors.

The results show that reduced models can preserve substantial distributional information, although performance varies across welfare concepts and information sets. For consumption, poverty status and position relative to the Gini-based inequality line can be predicted accurately with a small number of expenditure categories. Quintile classification is more demanding, especially when annual consumption is predicted from a single seasonal visit, but aggregated consumption categories still achieve high accuracy for seasonal welfare rankings. The most informative consumption predictors include food, meals consumed outside the home, roots and tubers, vegetables and rent, with some variation across poverty thresholds and seasons. For income, a small set of variables capturing crop sales, labour earnings, remittances, non-farm enterprise revenue, livestock sales, pensions and rental income accounts for much of the predictive information needed to classify individuals within the income distribution. Income poverty status reaches high accuracy with only a few predictors, while position relative to the Gini-based inequality line is captured particularly well by labour earnings.

Our analysis has two central implications. First, it shows that machine-learning-based feature selection can help identify a small set of welfare variables that retain much of the information needed for poverty and inequality monitoring. Second, it highlights that the optimal reduced module depends on the welfare concept being measured. Binary poverty and inequality-line classifications are easier to reproduce than quintile classification, and seasonal information matters for consumption-based welfare measurement. Reduced instruments should therefore be designed with explicit attention to the distributional outcome of interest, rather than treated as generic substitutes for full income or consumption surveys.



The remainder of the paper is organised as follows. Section~\ref{sec:ConceptualFram} sets out the welfare measurement framework used to construct the benchmark income and consumption aggregates. Section~\ref{sec:PreApproaches} reviews existing approaches to reduced income and consumption measurement. Section~\ref{sec:Methodology} presents the RF-RFE framework and defines the target classification outcomes. Section~\ref{sec:data} describes the Nigeria GHS-Panel data and the construction of the analytical sample. Section~\ref{sec:Results} presents the empirical findings for consumption and income classifications. Section~\ref{sec:Robust} reports robustness checks using alternative equivalence scales and machine-learning algorithms. Section~\ref{sec:Conclusion} concludes with implications for survey design and welfare monitoring.

\section{Welfare aggregates and the reduced-data measurement problem}
\label{sec:ConceptualFram}

This section sets out the welfare measurement framework used to construct the benchmark income and consumption aggregates against which the reduced-data models are evaluated. We focus on income and consumption expenditure, which capture related but distinct dimensions of household economic well-being. Income measures the flow of resources available to the household for current consumption and saving. Following the Canberra Group framework, household income includes monetary and in-kind receipts received at annual or more frequent intervals that are available for current consumption and do not reduce household net worth \citep{canberragroup2011}. This excludes asset sales, borrowing, windfall gains and capital transfers. Consumption expenditure, by contrast, measures the value of goods and services acquired or consumed by the household during the reference period \citep{DeatonZaidi2002}. In low- and middle-income settings, consumption is often preferred for poverty analysis because it is less volatile than income and may better approximate longer-term living standards, while income remains central for analysing earnings, transfers and inequality \citep{Deaton1997,Ravallion1998}.

In the empirical analysis discussed below, household income is defined as:
\begin{equation}
Y_h =
Y_h^{L}
+
Y_h^{A}
+
Y_h^{LV}
+
Y_h^{NF}
+
Y_h^{T}
+
Y_h^{O},
\end{equation}
where \(Y_h^{L}\) denotes labour income, \(Y_h^{A}\) crop income, \(Y_h^{LV}\) livestock income, \(Y_h^{NF}\) non-farm enterprise income, \(Y_h^{T}\) remittances and private or public transfers, and \(Y_h^{O}\) other income sources, including pensions, rental income, interest and dividends. Agricultural, livestock and non-farm enterprise income are measured net of production costs where the survey provides the relevant information, in line with standard LSMS practice for welfare measurement in low- and middle-income countries \citep{GroshGlewwe2000, CarlettoEtAl2015}.

Second, household consumption expenditure is defined as:
\begin{equation}
C_h =
C_h^{F}
+
C_h^{NF}
+
C_h^{H},
\end{equation}
where \(C_h^{F}\) denotes food consumption, \(C_h^{NF}\) non-food consumption expenditure, and \(C_h^{H}\) housing-related consumption, including actual rent for tenants and imputed rent for owner-occupiers and non-market tenants. Food consumption includes purchases, own production, gifts and food consumed outside the home. Non-food consumption excludes asset purchases, debt repayments, production inputs, transfers to other households and highly irregular ceremonial expenditures where their inclusion would introduce substantial noise into short-reference-period welfare measurement \citep{DeatonZaidi2002}. Social transfers in kind and the service flow from non-housing durable goods are not included in the benchmark aggregate because they cannot be consistently valued with the available data.

Both income and consumption components are annualised and aggregated at the household level. Because the distributional analysis is conducted at the individual level, household resources are assigned to household members. The baseline specification uses per capita income and consumption:
\begin{equation}
W_{ih} = \frac{W_h}{n_h},
\end{equation}
where \(W_h\) denotes either household income or household consumption and \(n_h\) is household size. This approach is transparent and widely used in poverty analysis, but it imposes the strong assumption that household needs rise proportionally with household size and that there are no economies of scale in consumption \citep{Deaton1997, Lanjouw1995}. For this reason, the robustness analysis also considers alternative equivalence scales. First, we use the square-root scale, which divides household resources by the square root of household size and therefore allows needs to increase less than proportionally with household size. Second, we use the modified OECD equivalence scale, which assigns a weight of 1 to the first adult, 0.5 to each additional adult and 0.3 to each child, thereby allowing for both economies of scale and differences in needs between adults and children \citep{hagenaars1994statistical}. These alternative scales reflect the long-standing recognition that welfare rankings and inequality estimates can be sensitive to assumptions about household sharing, demographic composition and economies of scale \citep{deaton1980economics, Lanjouw1995}.

\section{Existing approaches to reduced income and consumption measurement}
\label{sec:PreApproaches}

The measurement of income and consumption expenditure has conventionally relied on detailed survey modules covering multiple income sources, expenditure categories and recall periods. These instruments---often containing hundreds of items---remain the conventional benchmark for poverty and inequality analysis \citep{Deaton1997,DeatonZaidi2002}. However, their implementation is costly, time-consuming, and cognitively demanding for respondents. As a result, a growing literature has emerged proposing alternative approaches that approximate income or consumption expenditure using a reduced number of survey questions. This section discusses this strand of the literature.

\subsection{Asset-Based Wealth Indices}

One common approach replaces direct measurement of income or consumption with asset ownership and housing characteristics. The seminal work by \citet{FilmerPritchett2001} applied principal component analysis (PCA) to asset variables from Demographic and Health Surveys (DHS) to generate a wealth index closely correlated with consumption-based welfare rankings. The first principal component, interpreted as a latent index of long-run economic status, has since been widely used in contexts where income or expenditure data are unavailable \citep{montgomery2000measuring, mckenzie2005measuring, vyas2006constructing}.

Subsequent applications have extended this approach to classify households into national wealth quintiles. Tools such as the \textit{EquityTool} operationalize this approach by selecting a subset of asset variables that reproduces national quintile classifications with relatively high accuracy \citep{chakraborty2016equitytool}.\footnote{Chakraborty et al.\ (2016) report Cohen’s Kappa statistics generally falling between 0.60 and 0.75, indicating substantial agreement between the \textit{EquityTool}-based classifications and the full DHS wealth index, according to conventional benchmarks.} These instruments typically reduce 30–40 asset questions to approximately 10–15, thereby lowering survey burden while preserving relative ranking performance.

Asset-based indices offer some advantages. They are relatively stable over time, less affected by short-term income volatility, and easier for respondents to report accurately than detailed expenditure flows. However, empirical evidence shows that asset indices may perform well in ranking households at lowest extremes but exhibit weaker precision in the middle and upper tails of the distribution. Moreover, they also offer limited sensitivity to short-term welfare shocks \citep{howe2008measuring,harttgen2013comparison}, while cross-country and inter-temporal comparability can be affected by structural changes in asset ownership patterns, potentially leading to divergence between asset-based inequality measures and those derived from income or consumption aggregates \citep{smits2011towards, SahnStifel2003}. These considerations underscore that asset indices, while operationally attractive, may not fully approximate the distributional properties of consumption or income data used in standard poverty and inequality analysis.

\subsection{Proxy Means Tests and Poverty Scorecards}

A related but distinct literature develops proxy means tests (PMTs), which predict consumption or income using a small set of observable characteristics selected through supervised statistical models. Early work by \citet{GroshBaker1995} formalized regression-based targeting mechanisms that estimate household consumption as a function of demographic, educational, and housing variables. These models are typically calibrated on full consumption surveys and subsequently applied using a reduced questionnaire.

Poverty scorecards represent a simplified implementation of this logic. They use a small number (often 5–15) of observable indicators to estimate the probability that a household falls below a poverty line. Variables are selected to maximize predictive accuracy while minimizing interview time. Compared to PCA-based asset indices, PMTs are explicitly supervised and calibrated to a specific poverty threshold, making them more suitable for estimating poverty headcounts \citep{Schreiner2010}. Nevertheless, most PMT applications focus on binary poverty classification rather than approximating the consumption distribution. 

\subsection{Shortened Consumption Modules}

Rather than replacing consumption data with proxy indicators, another strand of the literature seeks to shorten expenditure modules themselves. Experimental evidence suggests that a relatively small subset of high-budget-share items captures a large share of total consumption variance. For example, \citet{Beegle2012} document how alternative questionnaire designs in Tanzania affect measurement accuracy and respondent burden. Similarly, \citet{GibsonKim2007} examine recall periods and item aggregation, highlighting trade-offs between precision and survey length.

These studies suggest that carefully selected consumption categories can reproduce aggregate poverty estimates with limited bias, although inequality measures are typically more sensitive to truncation and aggregation. The emphasis in this literature is on survey design features—recall periods, item aggregation, diary versus recall—rather than systematic selection of items. Nonetheless, the underlying objective is closely aligned with reducing the number of questions required to approximate welfare aggregates.

\subsection{Single-Question and Bracketed Income Measures}

In contexts where rapid data collection is essential—such as high-frequency phone surveys or labour force surveys—researchers often rely on a single self-reported income question. While appealing in its simplicity, this approach is known to suffer from substantial measurement error. Reviews by \citet{Moore2000} and \citet{Bound2001} document systematic under-reporting, non-response, and rounding (heaping) in income data. Moreover, \citet{Deaton1997} argues that income is generally more volatile and less reliably measured than consumption in developing-country settings.

To mitigate reporting error, some surveys employ bracketed income questions, asking respondents to select the interval within which their income falls.\footnote{For examples of bracketed income questions, see the University of Michigan’s Panel Study of Income Dynamics (\url{https://psidonline.isr.umich.edu}); the Federal Reserve Board’s Survey of Consumer Finances (\url{https://www.federalreserve.gov/econres/scfindex.htm}); and the U.S. Census Bureau’s American Community Survey (\url{https://www.census.gov/programs-surveys/acs/methodology.html}).} Evidence from unfolding bracket designs shows that this approach substantially reduces item non-response and improves response rates for sensitive financial questions \citep{Hurd2004,Kennickell1998}. However, while bracketing lowers respondent burden, it limits precision for distributional analysis, particularly when estimating inequality indices or poverty headcounts close to the threshold, as grouped income data can bias Gini coefficients downward and distort poverty measures when the poverty line falls within an interval \citep{CowellMehta1982,Ravallion1994}. Thus, while single-question or interval-based approaches may suffice for coarse ranking or trend monitoring, the literature suggests that they are generally inadequate for reconstructing detailed distributional statistics without strong parametric assumptions \citep{Kennickell1998,CowellVictoriaFeser1996}.

\subsection{Subjective Welfare Measures}

An alternative strategy elicits perceived economic status through subjective questions, such as self-assessed poverty or placement on an economic ladder. \citet{PradhanRavallion2000} and \citet{RavallionLokshin2002} report that subjective welfare measures correlate with consumption-based poverty status and can be used to derive \textit{implicit} poverty lines.

Although these questions require minimal survey time, they capture multidimensional perceptions of welfare rather than monetary consumption \textit{per se}. Cultural norms, reference groups, and reporting heterogeneity limit their comparability across contexts. As such, subjective measures are better interpreted as complements to, rather than substitutes for, consumption data in distributional analysis.

Taken together, the literature identifies several strategies for reducing survey length: constructing asset-based wealth indices through PCA; implementing supervised proxy means tests or poverty scorecards; shortening consumption modules by focusing on selected expenditure categories; relying on single or bracketed income questions; and eliciting subjective welfare assessments. Each approach involves trade-offs between survey cost, respondent burden and the precision with which the underlying welfare distribution can be recovered.

Most existing approaches are designed for specific purposes, typically poverty targeting, relative wealth ranking or rapid monitoring. They are less well suited to preserving several distributional properties simultaneously, including poverty status, quintile distribution and position relative to inequality-sensitive thresholds. This gap motivates the RF-RFE approach developed in the following section, which uses supervised feature selection to identify a parsimonious set of income and consumption variables that retains predictive information across multiple welfare classifications.

\section{Methodology}
\label{sec:Methodology}

Our primary objective is to identify a parsimonious subset of income and consumption components that can reproduce key distributional outcomes derived from the full welfare aggregates. The purpose is not only to reduce questionnaire length, but also to assess whether shorter survey instruments can preserve sufficient information for poverty and inequality analysis.

We organise the empirical exercise around three outcomes. 
First, we examine whether reduced survey instruments can predict poverty status under alternative poverty thresholds. Second, we evaluate whether they can identify individuals' position relative to the Gini-based inequality line. Third, we examine quintile classification as a broader test of whether reduced models preserve individuals' ranking across the welfare distribution. To implement these tasks, we use Random Forest Recursive Feature Elimination (RF-RFE), a supervised machine learning framework that combines ensemble prediction with iterative feature selection \citep{Breiman2001,Guyon2002,Hastie2009}.\footnote{For expositional simplicity, all equations are written without explicit sampling-weight notation. In the empirical implementation, however, the benchmark welfare distributions, poverty thresholds, quintile classifications, inequality lines and model-performance statistics are computed using the survey weights described in Section~\ref{sec:data}.}

\subsection{Random Forest Recursive Feature Elimination}
\label{sec:RF-RFE}

RF-RFE is a wrapper-based feature selection method that uses Random Forests as the base learner \citep{KohaviJohn1997,Guyon2002}. The algorithm begins with the full set of candidate predictors, estimates a predictive model, ranks variables according to their contribution to model performance, removes the least informative predictor, and repeats this process until a sequence of increasingly parsimonious models has been evaluated.

Let \(Y_i\) denote the target welfare outcome for individual \(i\). 

Depending on the empirical task, \(Y_i\) represents poverty status, position relative to the inequality line, or position in the quintile distribution. Let
\[
\mathbf{x}_i = (x_{i1},\dots,x_{ip})'
\]
denote the full set of \(p\) candidate predictors, including income sources, consumption categories and, where relevant, household characteristics. The objective is to identify a reduced subset of predictors \(S \subseteq \{1,\dots,p\}\) such that
\[
f(\mathbf{x}_{i,S}) \approx Y_i,
\qquad |S| \ll p,
\]
where \(\mathbf{x}_{i,S}\) denotes the vector of predictors for individual \(i\) restricted to the variables in \(S\), and \(f(\cdot)\) is the prediction function.

We split the data into an 80 percent training sample and a 20 percent held-out test sample. The split is implemented at the household level to avoid allocating members of the same household to both training and test samples. The cross-validation folds are also defined at the household level, ensuring that individuals from the same household are not assigned to both training and validation folds. All feature elimination, model selection and cross-validation procedures are conducted within the training sample. The test sample is reserved exclusively for final out-of-sample evaluation.

At the first step, the algorithm estimates a Random Forest using the full predictor set \(S_0=\{1,\dots,p\}\). Random Forests construct multiple decision trees on bootstrapped samples and aggregate their predictions, which improves predictive accuracy and reduces sensitivity to individual sample realisations \citep{Breiman2001}. For classification tasks, predictions are obtained by majority voting across trees. 

At each iteration, the algorithm computes variable-importance scores. In the baseline classification models, importance is measured using the mean decrease in Gini impurity, a standard variable-importance measure in Random Forest classification models \citep{Breiman2001}. 
The predictor with the lowest importance score is removed, the model is re-estimated on the reduced feature set, and the procedure continues until all nested subsets have been evaluated. The result is a learning curve that relates predictive performance to the number of retained variables.

To reduce dependence on a single training split, we embed the elimination procedure within repeated fivefold cross-validation, with \(K=5\) folds and \(R=5\) repetitions \citep{Hastie2009}. For a given subset \(S\), cross-validated risk is defined as:
\[
\widehat{R}_{CV}(S)
=
\frac{1}{R}
\sum_{r=1}^{R}
\frac{1}{n_{\text{train}}}
\sum_{\kappa=1}^{K}
\sum_{i\in \mathcal{D}^{(r)}_{\kappa}}
\mathcal{L}
\left(
Y_i,
f^{(-\kappa,r)}(\mathbf{x}_{i,S})
\right),
\]
where \(\mathcal{D}^{(r)}_{\kappa}\) denotes validation fold \(\kappa\) in repetition \(r\), \(f^{(-\kappa,r)}\) is the model trained on the remaining folds, \(n_{\text{train}}\) is the size of the training sample, and \(\mathcal{L}(\cdot)\) is the relevant loss function. Cross-validation ensures that feature rankings reflect stable predictive relationships rather than noise in a single training realisation. For classification tasks, we use zero-one loss:
\[
\mathcal{L}
\left(
Y_i,
f(\mathbf{x}_{i,S})
\right)
=
\mathbf{1}
\left\{
Y_i \neq f(\mathbf{x}_{i,S})
\right\}.
\]
Under this specification, empirical risk corresponds to the misclassification rate. For binary outcomes, such as poverty status or position relative to the inequality line, accuracy is defined as:
\[
\text{Accuracy}
=
\frac{TP+TN}{TP+TN+FP+FN},
\]
where \(TP\), \(TN\), \(FP\) and \(FN\) denote true positives, true negatives, false positives and false negatives. 

For quintile classification, the outcome has five categories. In this case, predictive performance is evaluated using multiclass accuracy, defined as the share of individuals correctly assigned to their observed quintile:
\[
\text{Accuracy}
=
\frac{1}{n}
\sum_{i=1}^{n}
\mathbf{1}
\left\{
\widehat{Q}_i = Q_i
\right\},
\]
where \(Q_i\) denotes the observed quintile of individual \(i\) and \(\widehat{Q}_i\) denotes the predicted quintile. 

Rather than selecting a single model solely on the basis of maximum accuracy, we report performance over the full sequence of nested models. This allows us to identify the smallest set of variables that reaches a high and stable level of predictive performance. This is important for survey design, where the relevant trade-off is not only predictive accuracy, but also the number of questions required to obtain that accuracy.

\subsection{Target variables}
\label{sec:targets}

The RF-RFE framework evaluates whether reduced survey instruments can reproduce three distributional outcomes derived from the full income and consumption aggregates: poverty status, position relative to the Gini-based inequality line, and quintile classification. These outcomes capture complementary dimensions of welfare measurement: identification of deprivation, classification relative to an inequality-sensitive threshold, and ranking across the full welfare distribution.

\subsubsection{Poverty status}

We begin with poverty status because it is the most common policy application of reduced welfare instruments. A poverty line defines the threshold separating poor from non-poor individuals. For consumption, we evaluate poverty status using both national and international poverty lines. The national poverty line reflects domestic welfare standards and cost-of-living conditions \citep{NBS2020}, while the international poverty line facilitates comparison with global poverty monitoring \citep{WorldBank2022}. For income, we focus on a relative poverty line, defined as 60 percent of median income, \(z^{R} = 0.6 \times \text{median}(y)\), following common practice in relative poverty measurement \citep{OECD2011,Eurostat2021}. For a given welfare aggregate \(y_i\), poverty status is defined as:
\[
P_i = \mathbf{1}\{y_i < z\},
\]
where \(z\) denotes the relevant poverty threshold. The reduced-data model is evaluated by its ability to reproduce this binary classification using a smaller set of income or consumption predictors.




\subsubsection{Inequality lines}

We also examine whether reduced models can identify whether individuals fall below or above the inequality line. Inequality lines define the level of welfare below which an additional transfer reduces inequality, and above which an additional transfer increases inequality \citep{LopezNoval2024}. They provide a distribution-sensitive threshold that differs from poverty lines, which are defined in relation to deprivation.

In this paper, we focus on the inequality line associated with the Gini index. Let \(F^{-1}(\cdot)\) denote the quantile function and let \(G\) denote the Gini coefficient. The Gini-based inequality line is given by:
\[
c_y =
F^{-1}
\left(
\frac{1+G}{2}
\right).
\]
We then define whether individual \(i\) lies above the inequality line as:
\(
R_i =
\mathbf{1}
\{y_i \geq c_y\}.
\)
The RF-RFE model is evaluated by its ability to reproduce this binary classification using a reduced set of predictors.

\subsubsection{Income and consumption quintiles}

Finally, we examine whether reduced instruments can reproduce individuals' broader position in the welfare distribution. We develop classification models with five categories corresponding to quintiles of the income or consumption distribution. Each quintile represents 20 percent of the population, ranked from the lowest to the highest level of welfare. Let \(y_i\) denote the welfare level of individual \(i\), where \(y_i\) may refer to income or consumption. Let
\[
F(y)=\frac{1}{n}\sum_{i=1}^{n}\mathbf{1}\{y_i\leq y\}
\]
denote the empirical distribution function. The \(k\)-th quintile threshold \(Q_k\) is defined as:
\[
Q_k
=
\inf
\left\{
y:
F(y)\geq \frac{k}{5}
\right\},
\qquad k=1,\dots,4.
\]
These thresholds partition the population into five mutually exclusive groups. The first quintile contains the poorest 20 percent of individuals, while the fifth quintile contains the richest 20 percent. The reduced-data model is evaluated by its ability to assign individuals to the same quintile as the full welfare aggregate.

\subsection{From reduced predictors to distributional outcomes}
\label{sec:reduced_outcomes}

The final step evaluates whether the reduced models preserve the distributional information required for poverty and inequality analysis. We compare predicted and observed classifications for poverty status, position relative to the Gini-based inequality line, and quintile classification. This allows us to assess whether reduced questionnaires can accurately classify individuals into policy-relevant welfare groups and whether a parsimonious set of income and consumption variables can retain enough information to support poverty and inequality monitoring.

Rather than selecting a single model solely on the basis of maximum accuracy, we report performance across the full sequence of nested models. This makes it possible to identify the smallest set of variables that reaches a high and stable level of predictive performance, which is the relevant trade-off for survey design.

\section{Data} 
\label{sec:data}

The empirical analysis uses the fourth wave of the Nigeria General Household Survey-Panel (GHS-Panel), collected in 2018/19 by the National Bureau of Statistics (NBS) of Nigeria in collaboration with the World Bank Living Standards Measurement Study--Integrated Surveys on Agriculture (LSMS-ISA) programme. The GHS-Panel is a nationally representative household panel survey designed to collect detailed information on agricultural production, household welfare, livelihoods and socio-economic conditions. The 2018/19 round is the fourth wave of the panel, following previous rounds conducted in 2010/11, 2012/13 and 2015/16 \citep{NBSGHSPanel2019}.

The survey is representative at the zonal level, covering Nigeria's six geopolitical zones: three in the North and three in the South. It also distinguishes between rural and urban areas, allowing the analysis to account for important spatial differences in livelihoods, welfare levels and access to markets and services. The analytical sample used in this paper contains 5,049 households with the information required to construct both income and consumption aggregates. All distributional estimates and model-performance statistics are computed using the GHS-Panel sampling weights. These weights account for the survey's stratified multi-stage design, differential probabilities of selection, non-response, and the combination of the refresh and long-panel samples. Because the analysis is conducted at the individual level, household welfare aggregates are assigned to household members and weighted to recover the population-representative individual welfare distribution. 

Wave 4 is particularly appropriate for this paper because it combines rich welfare information with detailed agricultural and seasonal data. Households were interviewed twice: first during the post-planting period, between July and September 2018, and then during the post-harvest period, between January and February 2019. This two-visit structure is important in the Nigerian context, where income, agricultural production and consumption can vary substantially across the agricultural calendar. The post-planting visit captures household conditions during a period in which many agricultural households face tighter liquidity constraints and rely on expected harvests, while the post-harvest visit captures realised production, crop sales and consumption decisions after harvest.

The GHS-Panel Wave 4 sample combines a refresh sample with a long-panel component. In Wave 4, 360 new enumeration areas were selected for the refresh sample, with 60 enumeration areas per geopolitical zone and 10 households selected in each enumeration area. This refresh sample was combined with a subsample of households from the original panel to preserve the longitudinal dimension of the survey. The combined Wave 4 sample consisted of 519 enumeration areas \citep{NBSGHSPanel2019}.

The survey includes three main instruments: a household questionnaire, an agriculture questionnaire and a community questionnaire. The household questionnaire provides information on demographics, education, health, labour, food and non-food expenditure, household non-farm enterprises, food security, shocks, safety nets, housing conditions, assets, information and communication technology, remittances and other sources of income. The agriculture questionnaire collects detailed information on land ownership and use, plot characteristics, household and hired farm labour, input use, crop production, harvest and disposition, livestock activities, agricultural capital, irrigation and fishing. The community questionnaire records information on local infrastructure, prices, labour conditions, land markets, community organisations, resource management and local events.

For income, the survey contains information on labour earnings, agricultural crop production, livestock activities, non-farm enterprises, remittances, transfers, pensions, rental income, savings and other income sources. For consumption, it includes detailed food expenditure, meals consumed outside the home, non-food expenditure, education, health and housing information. These modules allow us to construct full income and consumption aggregates that serve as benchmark welfare measures against which reduced-data models are evaluated.\footnote{For a more detailed discussion of the procedures used to construct the benchmark welfare aggregates, see Appendix~\ref{app:cons_inc_aggregate}.}

In this paper, the post-planting and post-harvest data are used both separately and jointly. This allows us to assess whether reduced survey instruments can recover welfare rankings, poverty status, position relative to the Gini-based inequality line and quintile classifications under different seasonal information sets. The working sample is restricted to households with the information required to construct the income and consumption aggregates and the corresponding distributional outcomes used in the RF-RFE analysis.

\section{Results} \label{sec:Results}

In this section, we present the primary findings regarding the modelling of the income and consumption distribution using a parsimonious set of variables. Specifically, we identify the key predictors determining individual positions within the distribution, focusing on quintile classification and status relative to both the poverty line and the inequality line. 

As detailed in Section \ref{sec:ConceptualFram}, baseline results are derived by adjusting household income and consumption by the number of household members to estimate per capita welfare aggregates. The sensitivity of our findings to this specification is addressed in the subsequent robustness section, where we apply alternative equivalence scales to individualise household income. To identify the most relevant predictors, we implemented RF-RFE using a random forest base learner \citep{Breiman2001}, with model performance validated through repeated five-fold cross-validation.

\subsection{Predicting consumption}
As detailed in Table \ref{tab:food_items} in the Appendix, the survey instrument captures household consumption across more than 150 food items, nine categories for meals consumed outside the home, approximately 90 non-food categories, and various expenditures on education and health. This high-dimensional data, comprising over 240 individual items, presents significant computational challenges for the implementation of RF-RFE. To address this complexity, we employ a two-tiered aggregation strategy to streamline the feature space while maintaining economic relevance.

The first level of aggregation reduces the data to 36 categories. This includes 15 food categories and four categories for meals consumed outside the home, as specified in Table \ref{tab:food_items}. Non-food items are consolidated into 13 categories following the structure in Table \ref{tab:nonfood}, alongside a dedicated category for paid or imputed rent. Finally, education expenditures are grouped into one category and two for health, distinguishing between outpatient costs (consultations, transport, and drugs) and hospital stays.

In instances where this granular level of aggregation fails to yield sufficient predictive accuracy, we implement a more parsimonious model. This second tier further aggregates sub-categories into broader headings, resulting in 18 consumption categories. In this configuration, all food items, all meals consumed outside the home, and all health-related expenditures are consolidated into their respective single variables to optimize model performance and interpretability. 

This aggregation strategy is consistent with evidence from survey experiments showing that the design of consumption modules, including item aggregation and recall structure, can substantially affect measured consumption and poverty estimates \citep{BeegleEtAl2012}. It also reflects the practical trade-off emphasised in the welfare-measurement literature: detailed expenditure modules improve coverage but increase respondent burden and may introduce reporting noise, while broader categories can preserve much of the relevant welfare information at lower cost \citep{DeatonZaidi2002}.

\begin{figure}[htbp]
\centering
\begin{tabular}{ll}
(a) Aggregated categories & (b) Disaggregated categories \\
\includegraphics[width=0.415\textwidth]{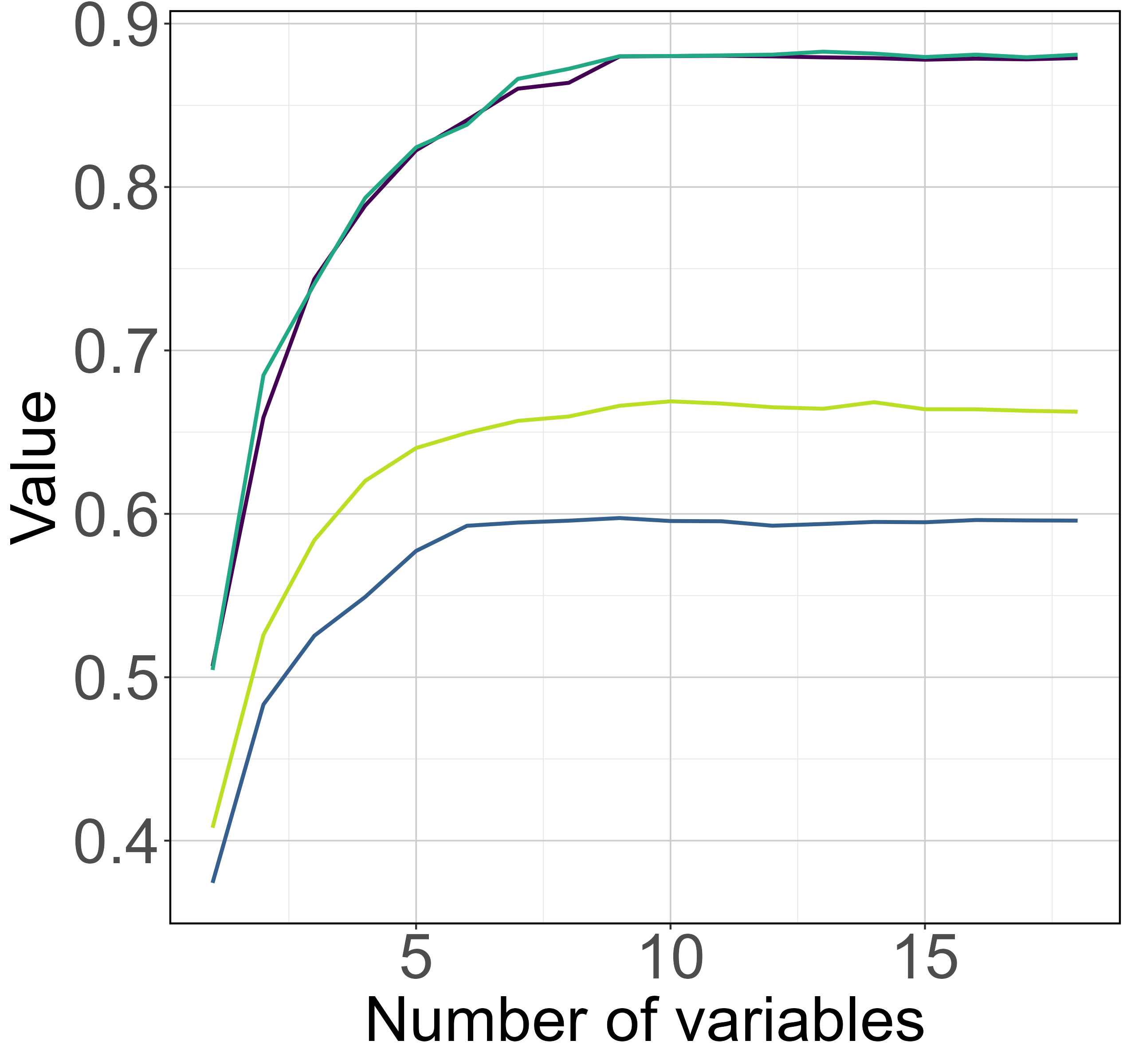} &
\includegraphics[width=0.59\textwidth]{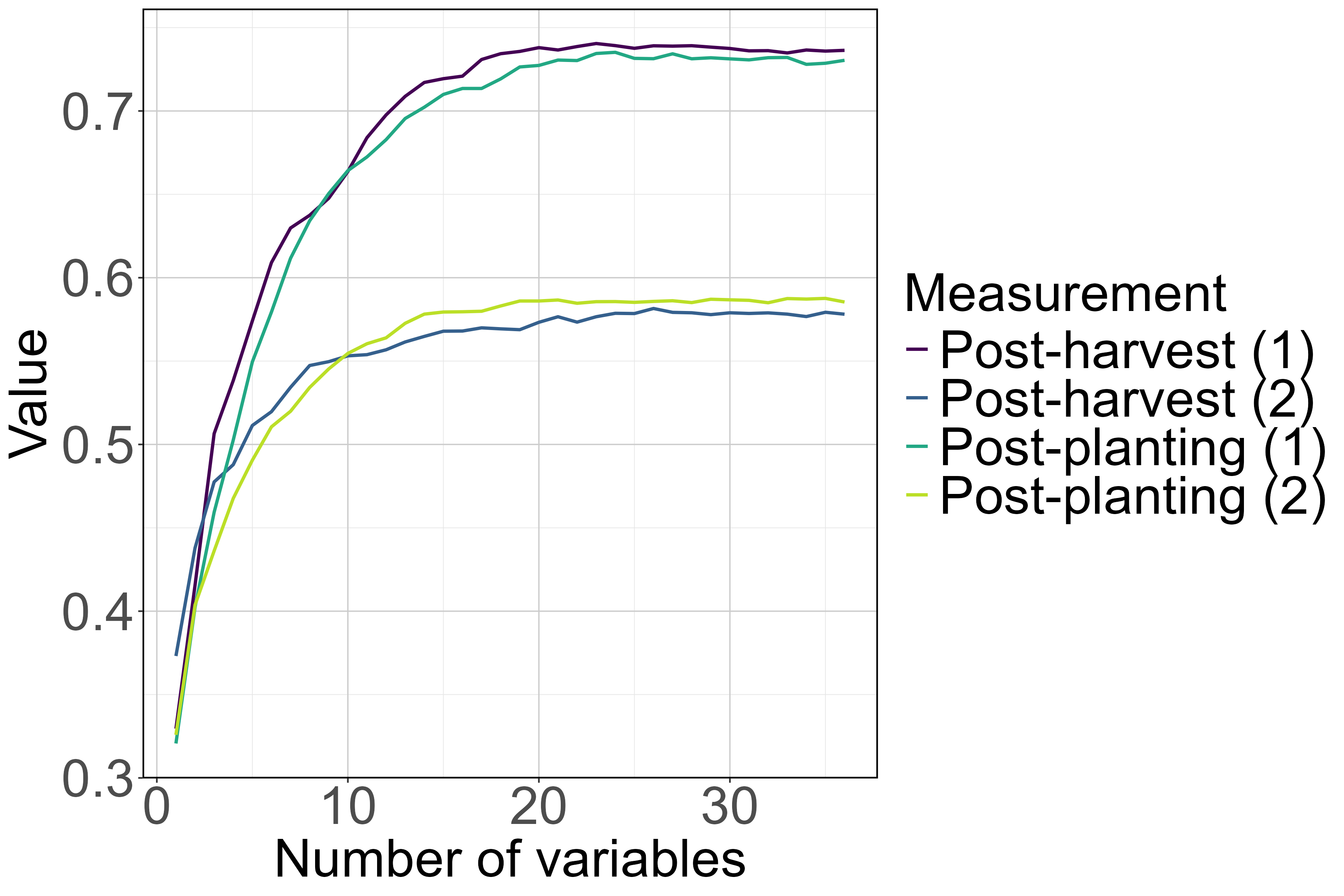} \\
\end{tabular}
\caption{Model accuracy in predicting consumption quintiles by number of variables}
\label{fig:acc_quintile}
\begin{flushleft}
\footnotesize \textit{Source:} Authors' compilation.
\end{flushleft}
\end{figure}

We first focus on predicting the position of individuals within the quintiles of the consumption distribution. Figure \ref{fig:acc_quintile} illustrates the accuracy of the models as a function of the number of variables, using both PP and PH data. For each dataset, two  models are estimated: (1) represents the prediction of the consumption distribution specifically for the period corresponding to the data collection, while (2) is used for models that rely on these seasonal datasets to predict aggregate annual consumption.

The right panel illustrates the results obtained using the 36 consumption categories. These findings suggest that with this level of  aggregation, the maximum accuracy for the seasonal models, denoted as (1), is approximately 73 percent. In contrast, when predicting annual consumption (Model 2), precision declines significantly to 54 percent, remaining relatively stagnant even as the number of features increases. When using a less granular set of consumption categories (left panel of Figure \ref{fig:acc_quintile}), predictive accuracy for seasonal data improves markedly, exceeding 80 percent with as few as five variables for both the post-harvest and post-planting periods. However, identifying an individual's quintile within the annual consumption distribution remains considerably more challenging. Accuracy peaks at approximately 65 percent for post-planting data and 60 percent for post-harvest data. 

The stronger performance of the aggregated specification suggests that predictive accuracy does not necessarily increase with more granular expenditure information. In this setting, broader consumption categories appear to reduce idiosyncratic item-level variation while retaining the budget components that most strongly differentiate households across the welfare distribution. This finding is consistent with the survey-design literature, which shows that shorter or more aggregated consumption modules can reproduce broad welfare patterns even though they may be less suitable for precise item-level expenditure analysis \citep{BeegleEtAl2012,GibsonKim2007}.

\begin{figure}[tbhp]
\centering
\begin{tabular}{ll}
(a) Aggregated categories & (b) Disaggregated categories \\
\includegraphics[width=0.37\textwidth]{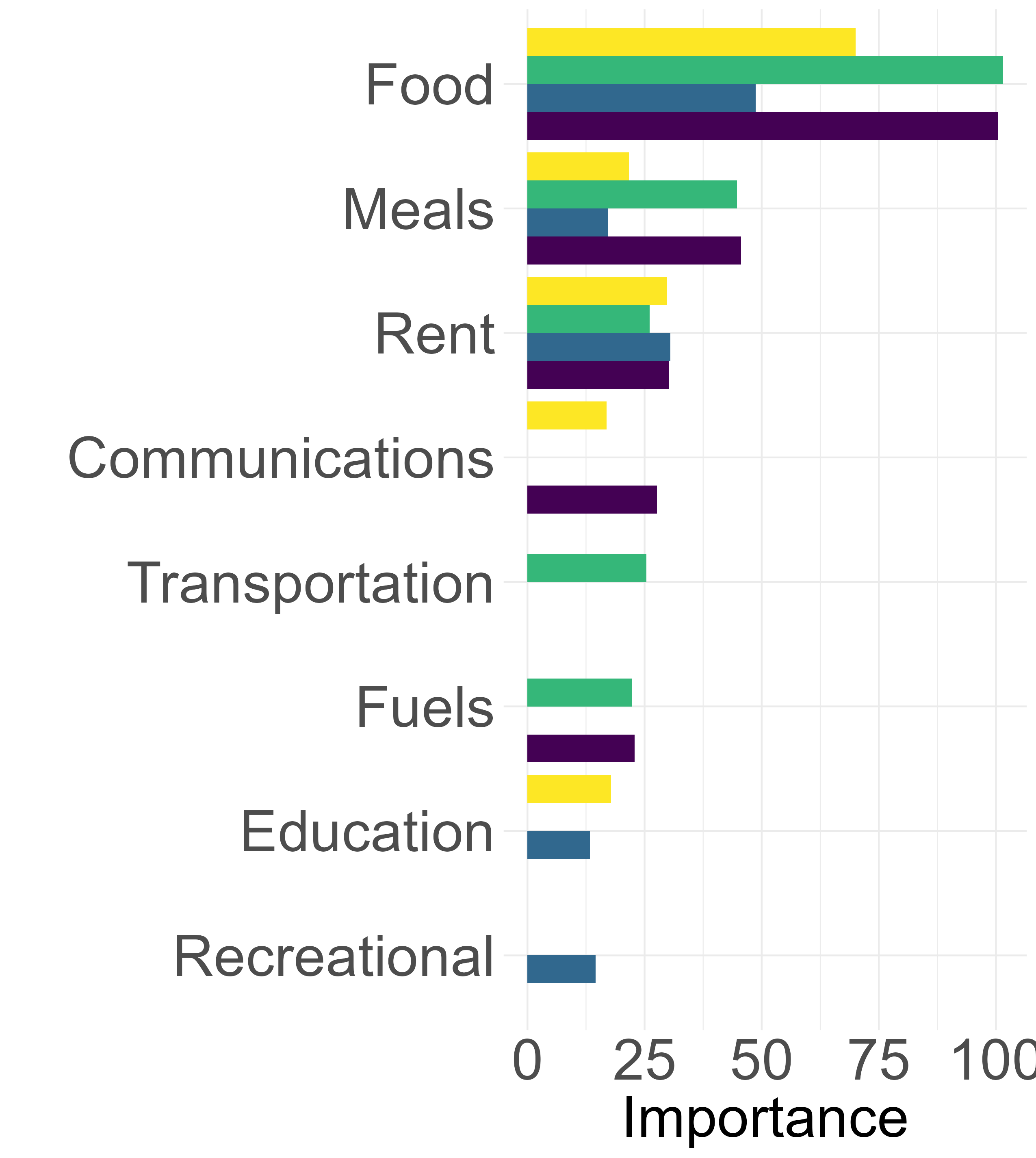} &
\includegraphics[width=0.6\textwidth]{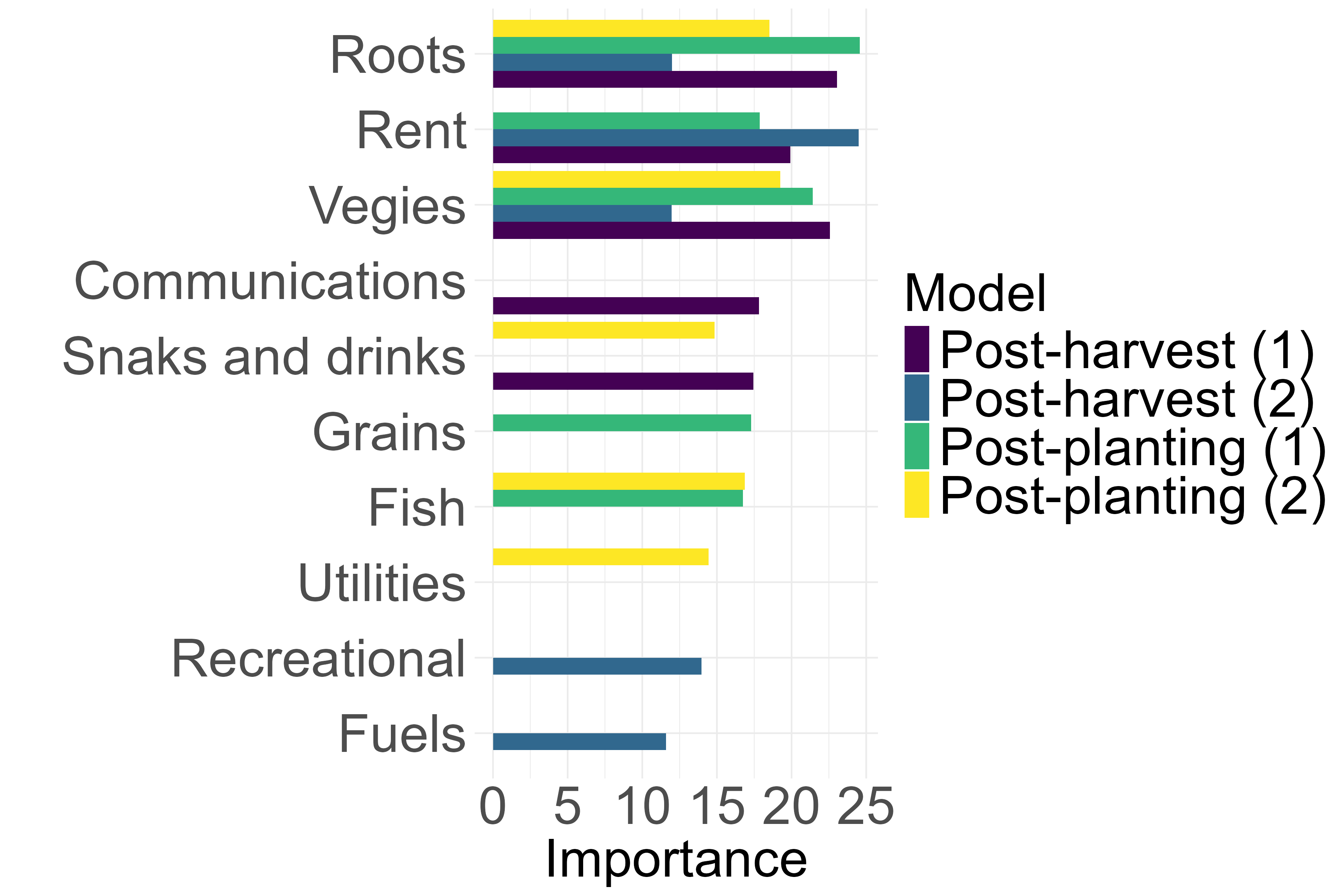} \\
\end{tabular}
\caption{Five most relevant variables to predict consumption quintiles using post-planting and post-harvest data}
\label{fig:varimp_quintile}
\begin{flushleft}
\footnotesize \textit{Source:} Authors' compilation.
\end{flushleft}
\end{figure}

We now analyse the primary predictors of consumption quintile classification. Figure \ref{fig:varimp_quintile} displays the five most influential categories for the four models considered across both levels of aggregation. In panel (a), \textit{food}, \textit{meals (outside the home)} and \textit{rent} emerge as the overwhelmingly dominant predictors across all seasonal and annual specifications. The importance of the aggregated food category is nearly double that of the next most significant variable. In the disaggregated model (Panel b), the predictive power is more evenly distributed across a wider set of categories. \textit{roots}, \textit{rent}, and \textit{vegetables} stand out as the primary pillars of the model. Interestingly, \textit{grains} and \textit{fish} appear as critical predictors specifically for the post-planting (lean season) models. A significant observation across both panels is the role of  \textit{rent}. Its importance is consistently high and exhibits less seasonal fluctuation than food-related items, suggesting that housing expenditure provides a relatively stable indicator of longer-term consumption levels.

The prominence of food expenditure is consistent with the central role of food in consumption-based welfare measurement in low- and middle-income settings, where food remains a large budget component and is closely linked to material deprivation \citep{Deaton1997}. The importance of rent is also consistent with standard welfare-aggregate practice, which treats housing services as a key component of household consumption and imputes rents for owner-occupiers and non-market tenants when observed rental payments are unavailable \citep{DeatonZaidi2002}.

\begin{figure}[htbp]
\centering
\begin{tabular}{ll}
(a) International poverty line & (b) National poverty line \\
\includegraphics[width=0.415\textwidth]{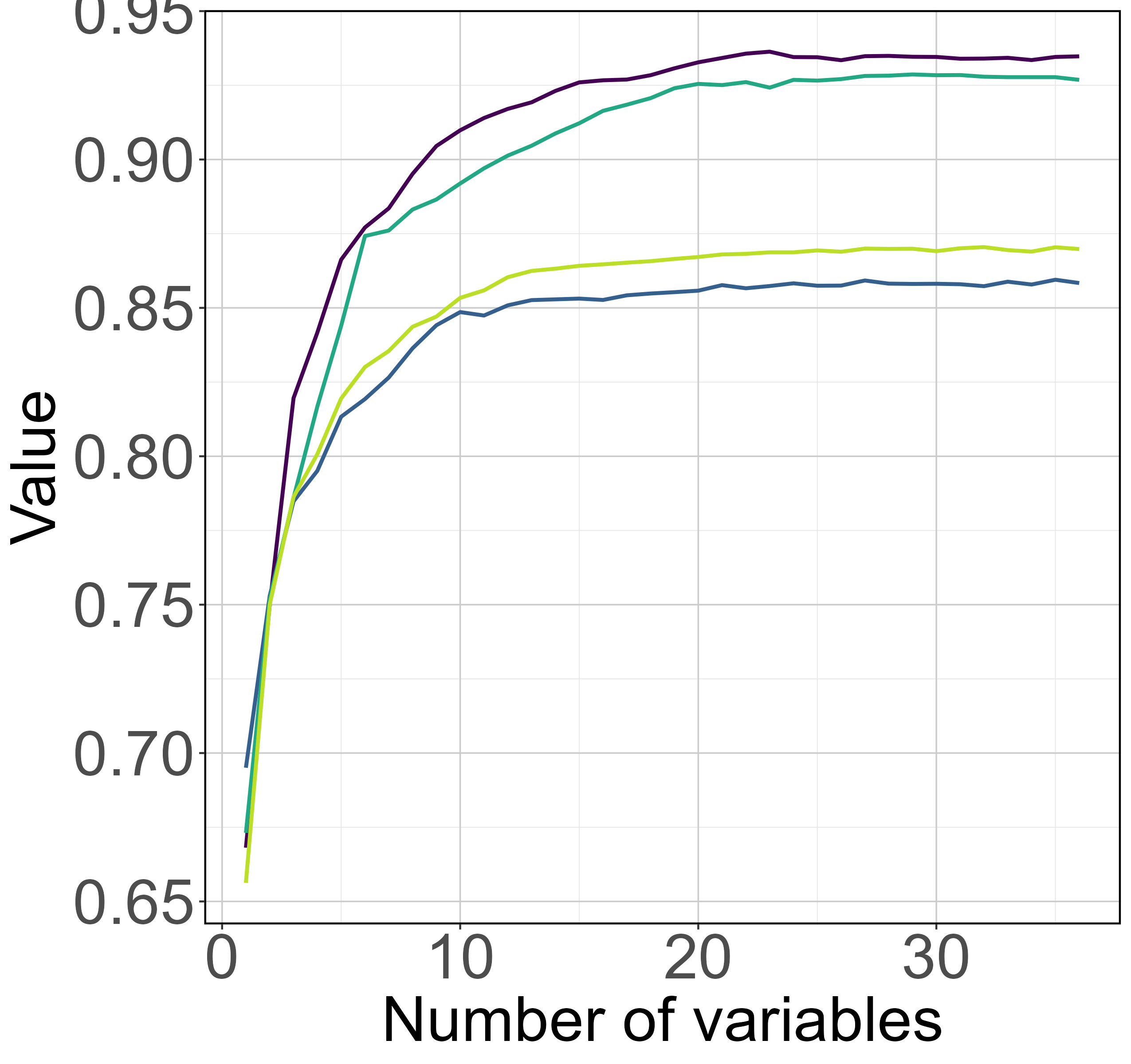} &
\includegraphics[width=0.6\textwidth]{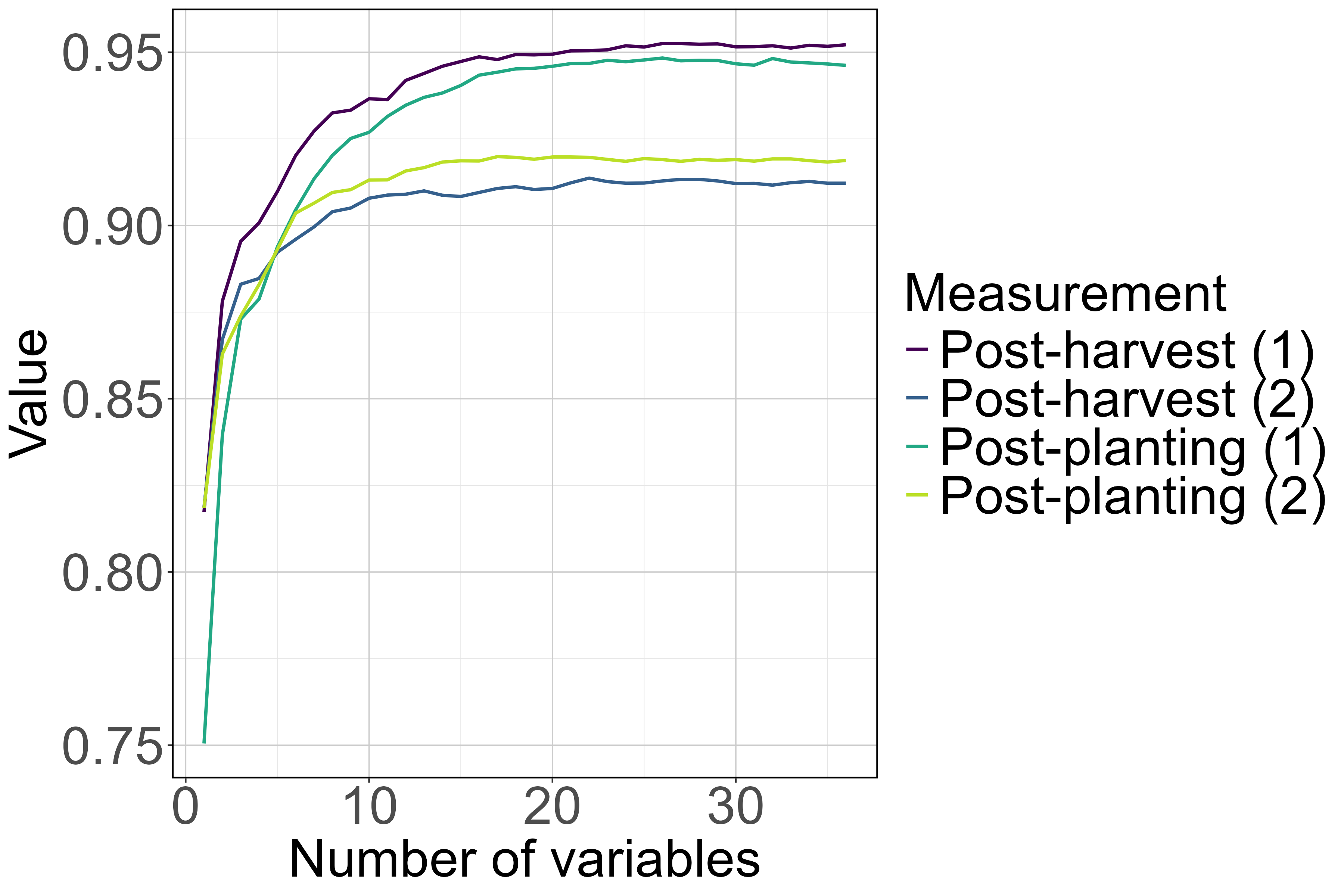} \\
\end{tabular}
\caption{Model accuracy in predicting poverty status by number of variables}
\label{fig:acc_poverty}
\begin{flushleft}
\footnotesize \textit{Source:} Authors' compilation.
\end{flushleft}
\end{figure}

We now turn our attention to the prediction of poverty status. Figure \ref{fig:acc_poverty} illustrates the accuracy of models derived from post-planting and post-harvest data, comparing predictions for seasonal versus annual consumption. The left panel presents the results for the international poverty line. Post-planting and post-harvest datasets demonstrate comparable performance in identifying poverty status, consistently surpassing the 80 percent accuracy threshold for both annual and seasonal consumption models. Specifically, the annual consumption model reaches this benchmark using only three variables, whereas the seasonal consumption model requires five variables to achieve an accuracy exceeding 80 percent. These results suggest that identifying poverty status (a binary classification) is considerably more straightforward than assigning individuals to specific consumption quintiles.

This pattern is consistent with the broader targeting literature. Binary poverty classification is generally less demanding than recovering a household's location across the full welfare distribution, which is why proxy means tests and poverty scorecards are typically designed around poverty status rather than full distributional reconstruction \citep{GroshBaker1995,CoadyGroshHoddinott2004}. The present results extend this logic by showing that a small set of consumption categories can perform well not only for poverty identification but also, with lower accuracy, for broader welfare ranking.

Regarding the national poverty line, the results in the right panel suggest that high predictive precision is attainable with minimal features. Specifically, the models achieve over 80 percent accuracy using only two variables across nearly all specifications. The sole exception is the model predicting seasonal poverty status based on post-planting data, which requires three variables to reach this same threshold of accuracy.

\begin{figure}[htbp]
\centering
\begin{tabular}{ll}
(a) International poverty line & (b) National poverty line \\
\includegraphics[width=0.37\textwidth]{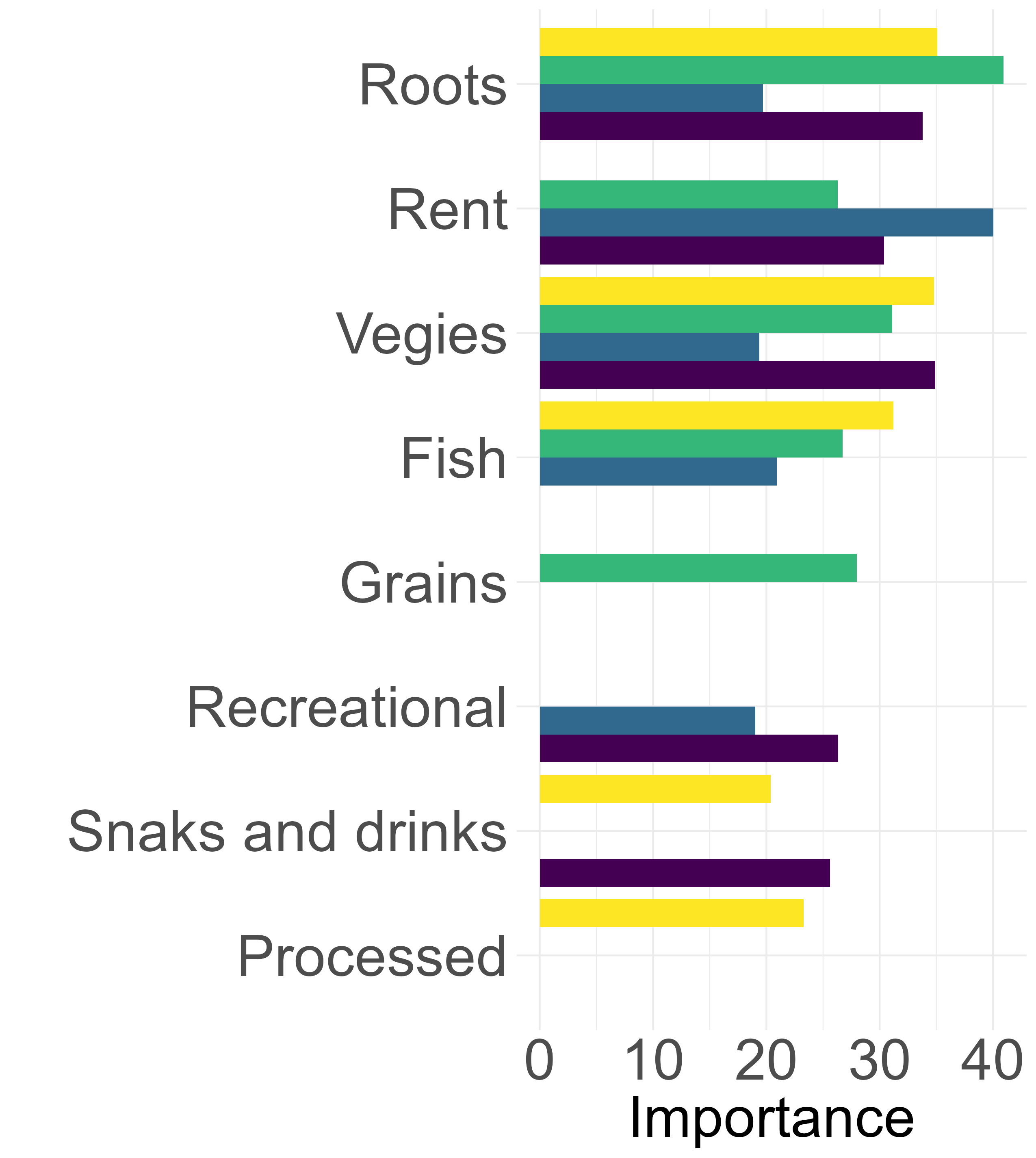} &
\includegraphics[width=0.6\textwidth]{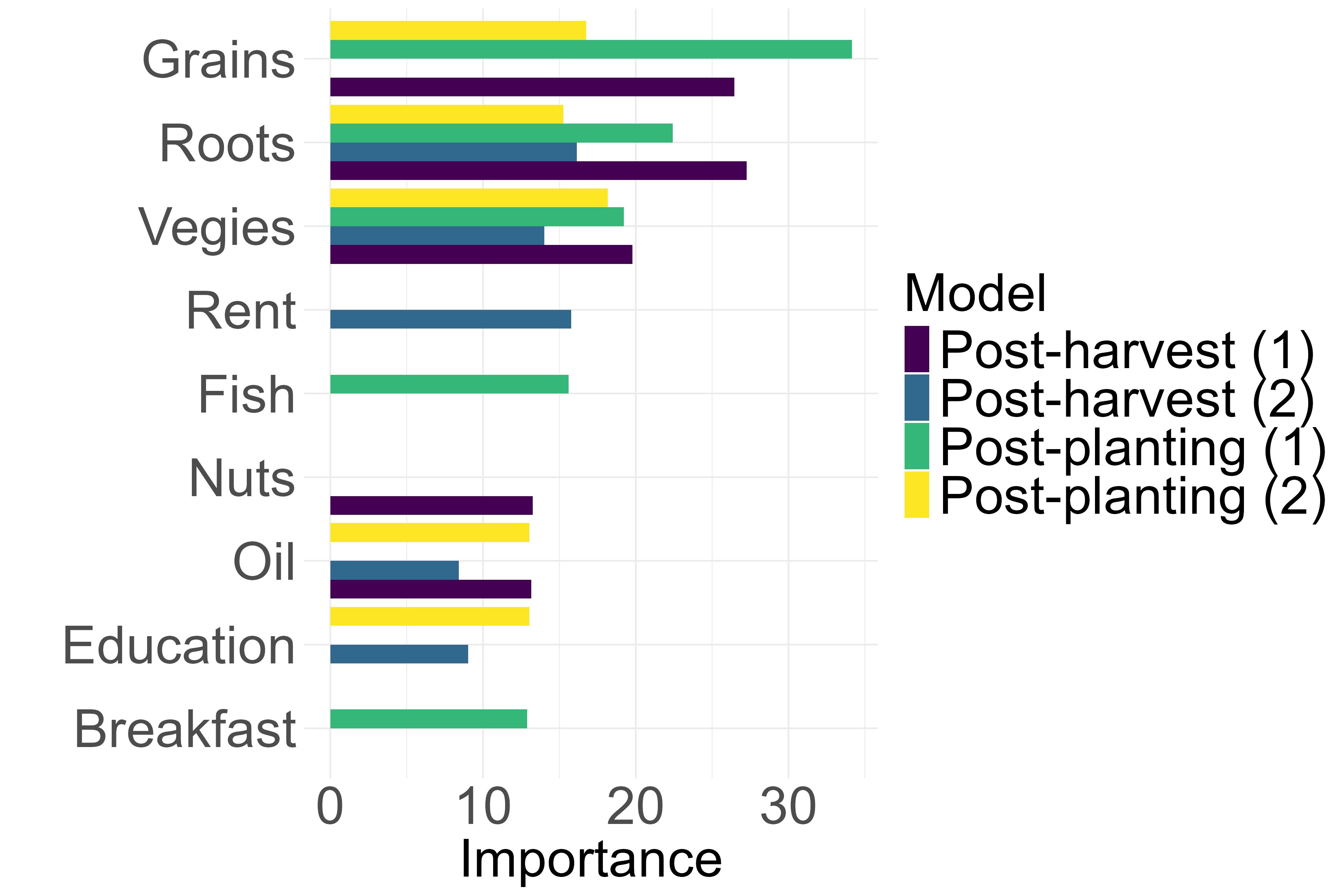} \\
\end{tabular}
\caption{Five most relevant variables to predict poverty using post-planting and post-harvest data}
\label{fig:varimp_poverty}
\begin{flushleft}
\footnotesize \textit{Source:} Authors' compilation.
\end{flushleft}
\end{figure}

The primary predictors of poverty status, illustrated in Figure \ref{fig:varimp_poverty}, reveal that the consumption profiles associated with poverty are highly dependent on both seasonal economic conditions and the specific poverty threshold employed. Several clear patterns emerge from this comparison.

First, \textit{food} staples consistently dominate the identification of poverty under both definitions. \textit{Roots} and \textit{vegetables} rank among the most influential predictors across nearly all specifications, while grains become particularly prominent under the national poverty line. The consistent importance of these categories underscores that staple food consumption remains the core dimension distinguishing poorer households from those with higher welfare levels.

Second, seasonality significantly affects which expenditures gain predictive relevance. Under the international poverty line, post-harvest models assign greater importance to discretionary and higher-value items such as recreation, snacks, drinks, and processed foods. This may be consistent with relaxed liquidity constraints following crop sales. In contrast, post-planting models, which correspond to the ``lean season'', place more weight on essential consumption such as fish, reflecting tighter budgets and a shift toward subsistence-oriented spending.

These seasonal shifts are consistent with evidence that agricultural households in developing countries face strong seasonal variation in income, liquidity and consumption opportunities. Existing studies show that the extent to which consumption tracks seasonal income depends on households' ability to smooth resources over the agricultural cycle and on their exposure to vulnerability and liquidity constraints \citep{Paxson1993,DerconKrishnan2000}. In the Nigerian context, the difference between post-planting and post-harvest predictors therefore reinforces the importance of collecting or modelling seasonal information when using reduced survey instruments.

Finally, the national poverty line places a stronger emphasis on education-related expenditures. Compared with the international line, housing-related expenditure (rent) becomes less dominant, and the predictor set shifts toward a more detailed food basket and basic welfare-related spending.

\begin{figure}[htbp]
\centering
\includegraphics[width=0.7\textwidth]{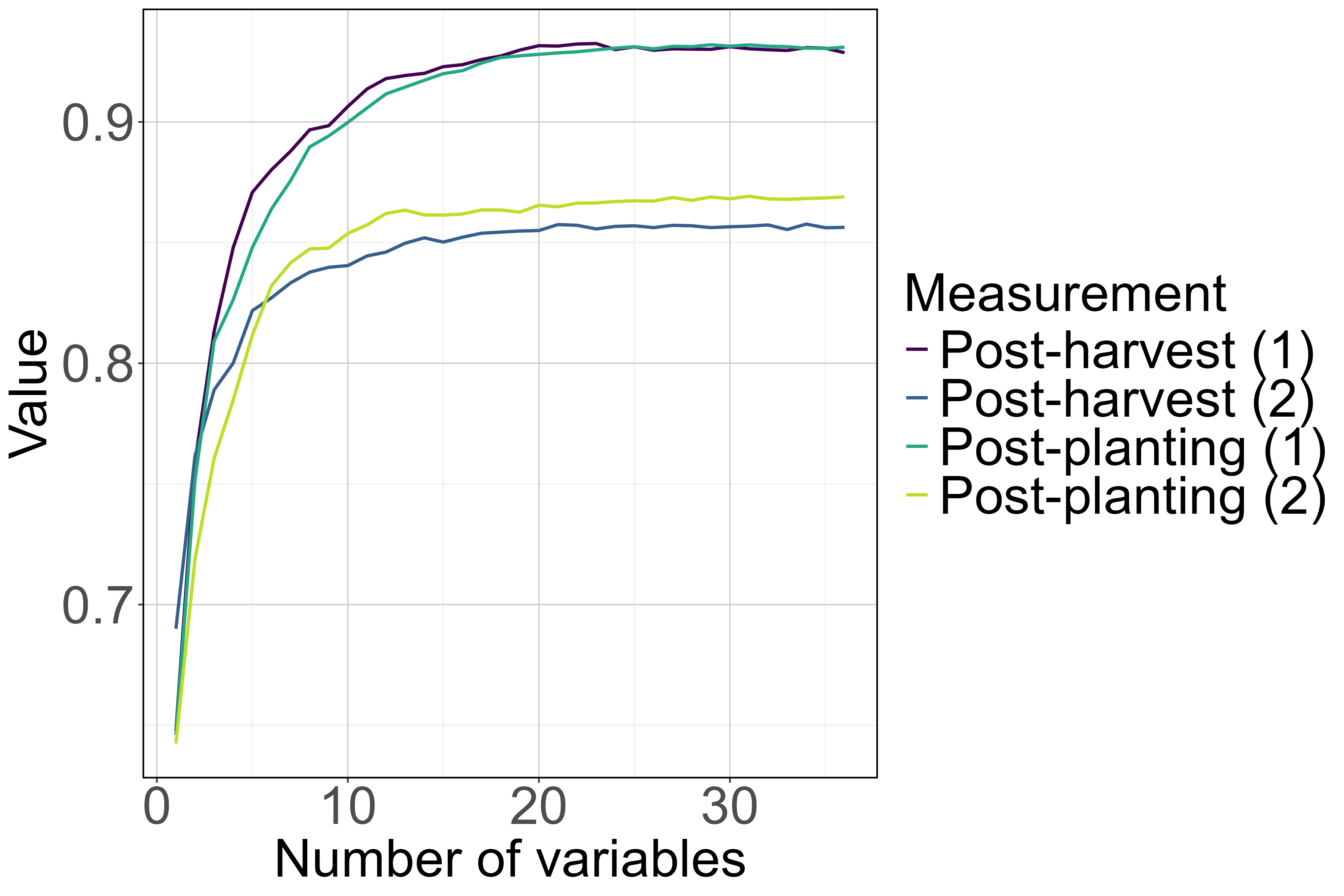}
\caption{Model accuracy in predicting individuals' position relative to the inequality line}
\label{fig:acc_line}
\begin{flushleft}
\footnotesize \textit{Source:} Authors' compilation.
\end{flushleft}
\end{figure}

Turning our attention to inequality lines, we explore the predictive accuracy of the models in identifying whether individuals fall above or below this specific threshold. As illustrated in Figure \ref{fig:acc_line}, and consistent with our findings regarding the poverty lines, high levels of accuracy are achieved using a limited number of variables. Specifically, when predicting seasonal consumption positions, the models require only three variables to surpass the 80 percent accuracy threshold. In contrast, for annual consumption predictions, five variables are necessary to achieve a comparable level of precision.

\begin{figure}[htbp]
\centering
\includegraphics[width=0.7\textwidth]{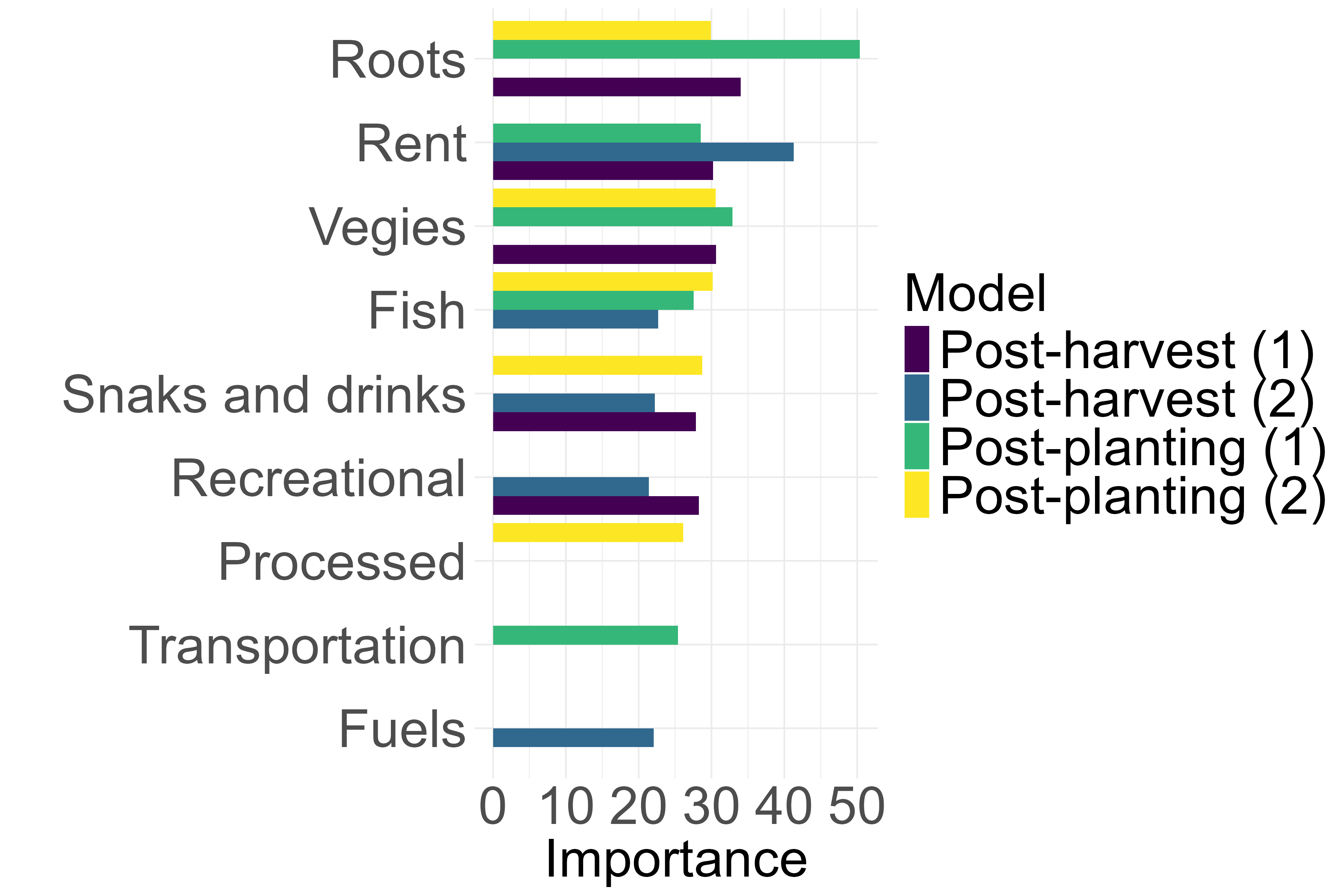}
\caption{Five most relevant variables for predicting individuals' position relative to the inequality line based on post-planting and post-harvest data}
\label{fig:varimp_line}
\begin{flushleft}
\footnotesize \textit{Source:} Authors' compilation.
\end{flushleft}
\end{figure}

The variable importance rankings for the inequality line (Figure \ref{fig:varimp_line}) illustrate a shift toward housing, infrastructure, and consistent discretionary spending. \textit{Roots} and \textit{rent} stand out as the most dominant features, with rent showing a particularly high importance in the post-harvest annual model (Model 2). This indicates that housing costs are a primary structural differentiator for the upper end of the distribution.

The data also reveals the importance of lifestyle and mobility indicators. Transport and fuels appear as significant predictors, categories that typically require higher levels of disposable income. Furthermore, discretionary items like snacks and drinks, recreational expenses, and processed foods maintain high importance across both the post-planting and post-harvest periods. This suggests that, unlike the seasonal patterns observed for poverty, these expenditures remain a persistent feature of household budgets for those above this threshold.

This result is consistent with the interpretation of inequality lines as distribution-sensitive thresholds rather than deprivation thresholds. Unlike poverty lines, which identify minimum welfare shortfalls, inequality lines identify the point in the distribution at which additional income or consumption changes the level of inequality \citep{Roope2021,LopezNoval2024}. It is therefore unsurprising that variables associated with more stable housing conditions, mobility and discretionary expenditure become more informative than the basic consumption items that dominate poverty classification.

\subsection{Predicting income}

We now turn to predicting income groups using a parsimonious set of variables. Following the procedure outlined in Appendix~\ref{app:income_data} to estimate income aggregates, we identify 36 potential predictors of an individual's position along the income distribution. Half of these variables represent positive income streams, while the remaining half comprises agricultural and production costs that enter negatively into the household income calculation.

\begin{figure}[htbp]
\centering
\begin{tabular}{ll}
(a) Accuracy & (b) Variables \\
\includegraphics[width=0.49\textwidth]{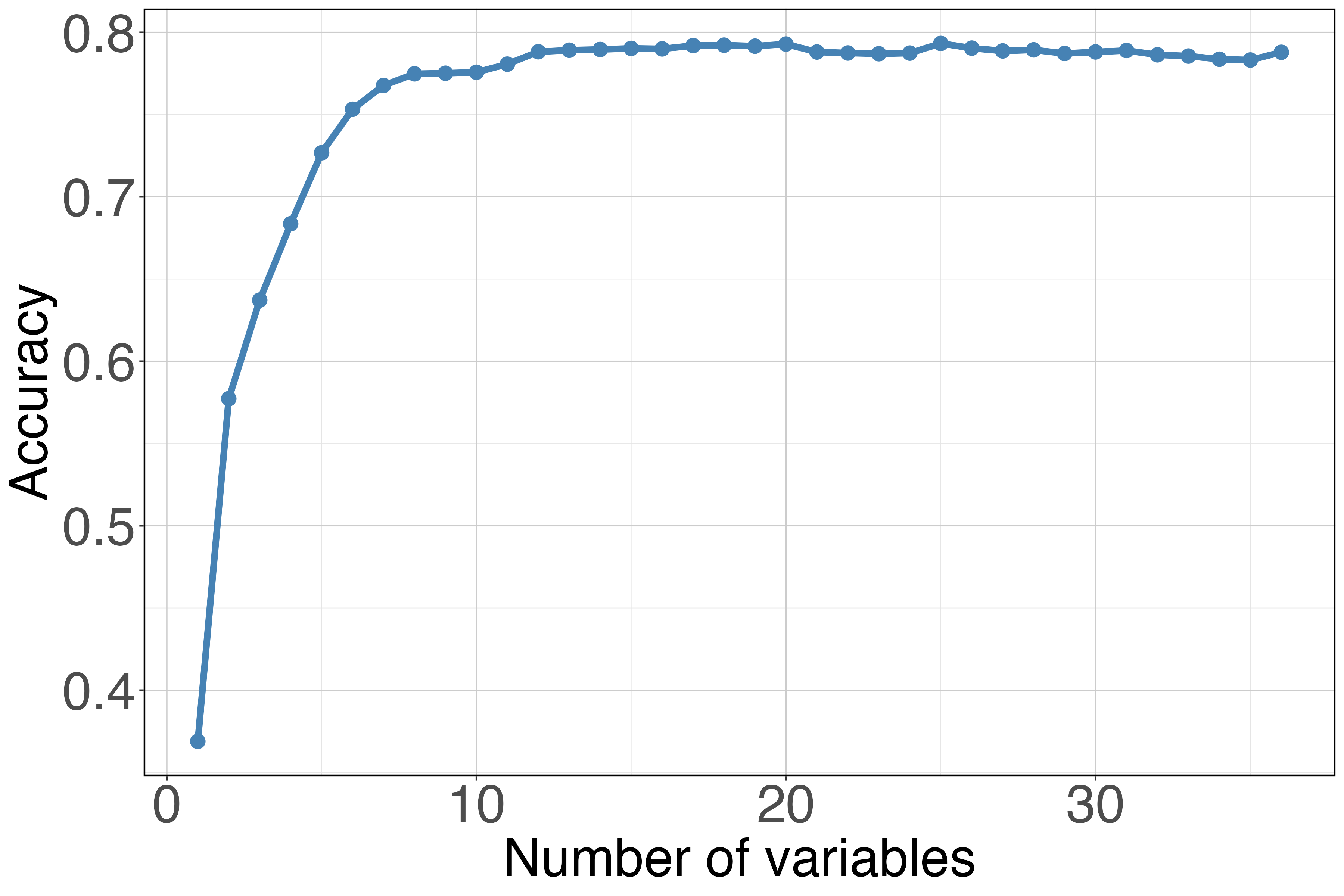} &
\includegraphics[width=0.49\textwidth]{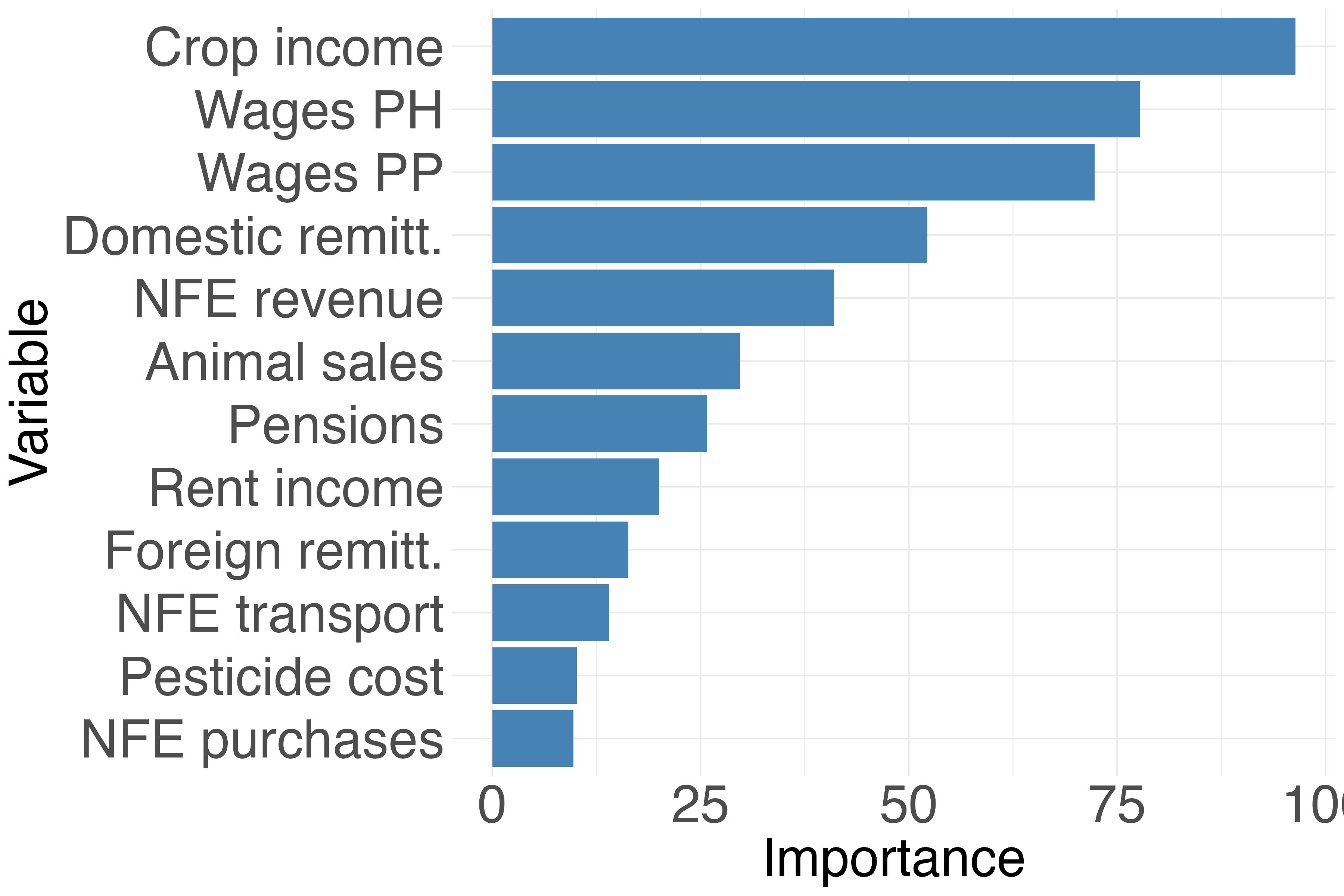} \\
\end{tabular}
\caption{Model accuracy and most relevant variables in predicting income quintiles}
\label{fig:income_quintile}
\begin{flushleft}
\footnotesize \textit{Source:} Authors' compilation.
\end{flushleft}
\end{figure}

\begin{figure}[tbph]
\centering
\begin{tabular}{ll}
(a) Accuracy & (b) Variables \\
\includegraphics[width=0.49\textwidth]{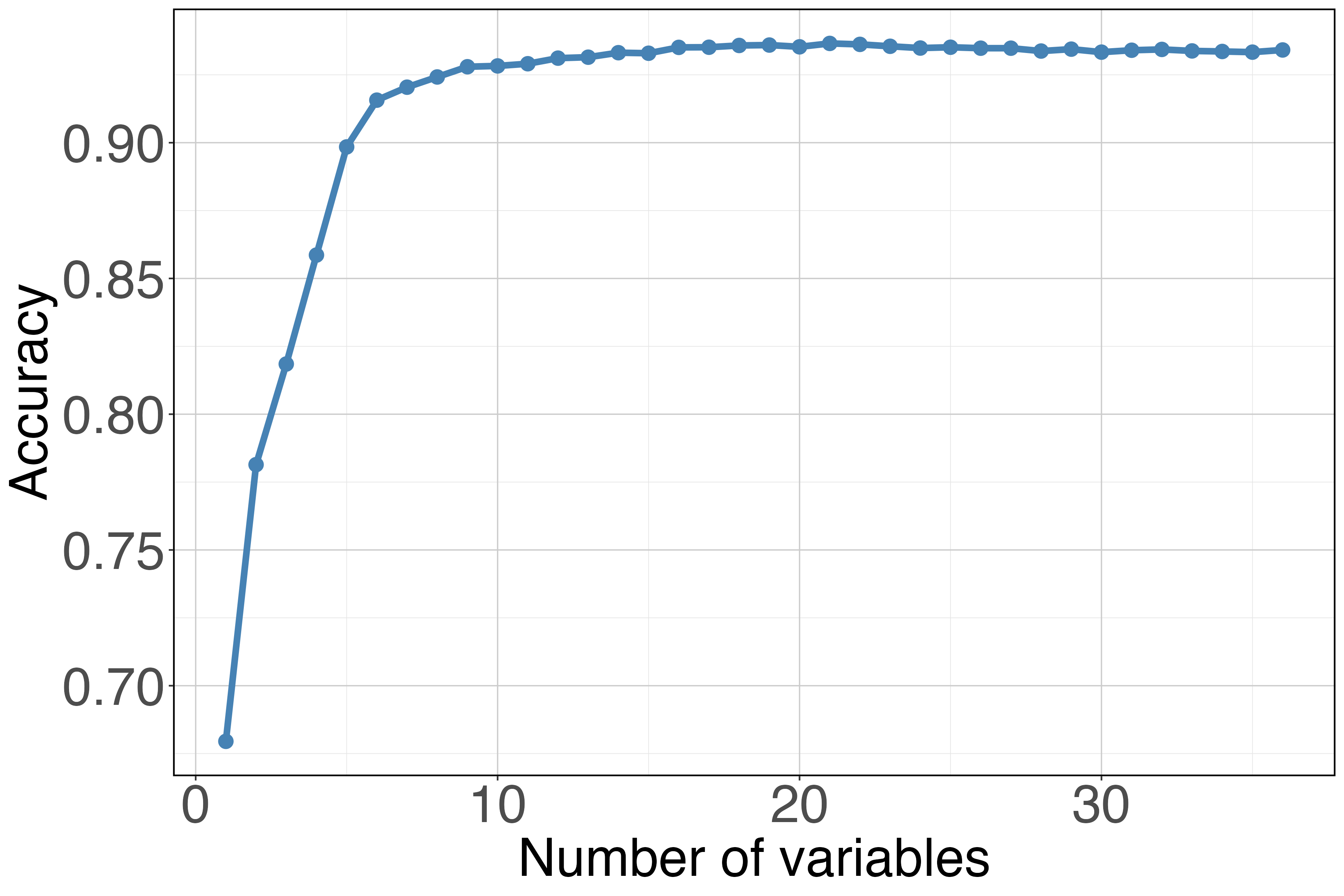} &
\includegraphics[width=0.49\textwidth]{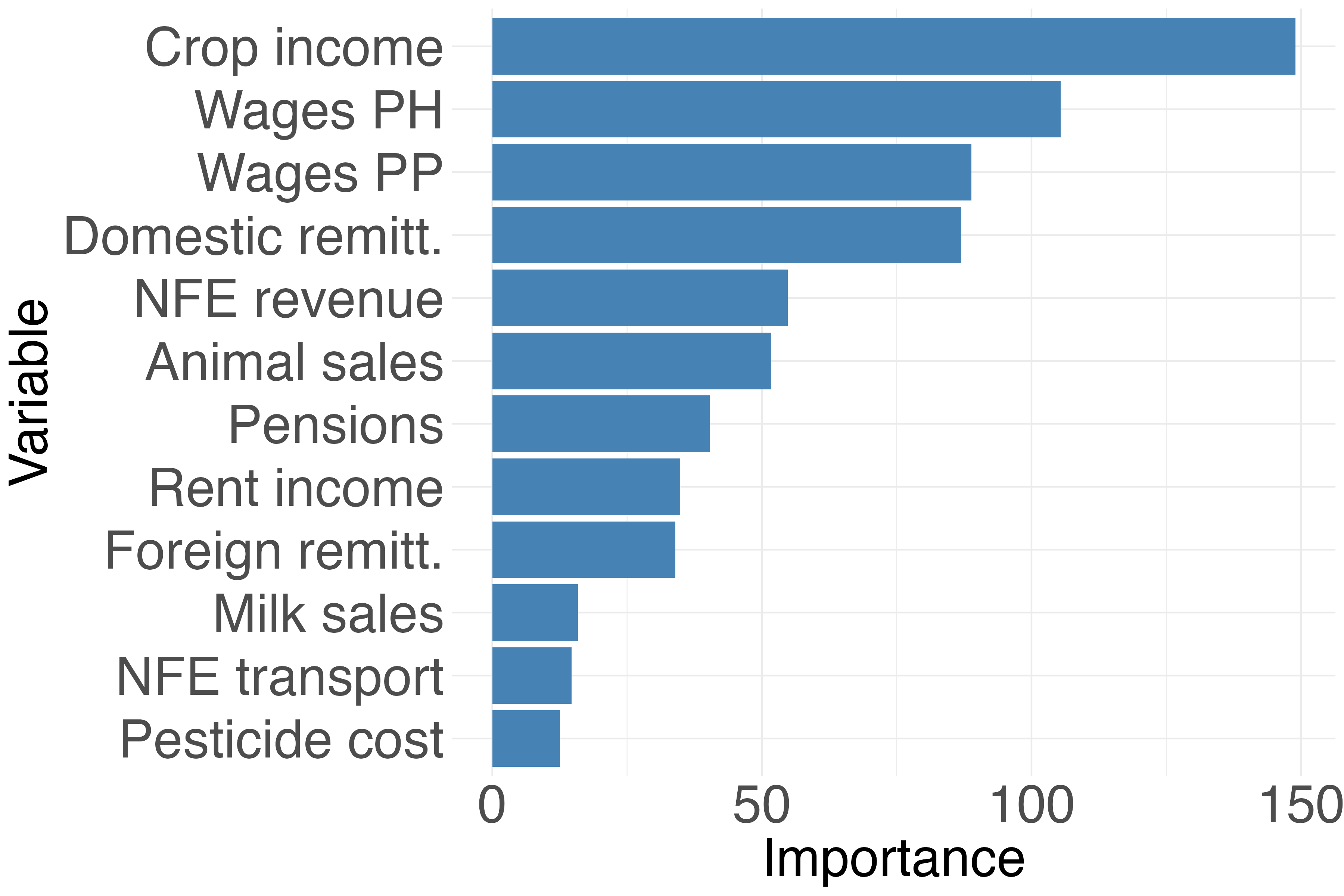} \\
\end{tabular}
\caption{Model accuracy and most relevant variables in predicting individuals' income position relative to the national poverty line}
\label{fig:income_NLP}
\begin{flushleft}
\footnotesize \textit{Source:} Authors' compilation.
\end{flushleft}
\end{figure}

\begin{figure}[htbp]
\centering
\begin{tabular}{ll}
(a) Accuracy & (b) Variables \\
\includegraphics[width=0.49\textwidth]{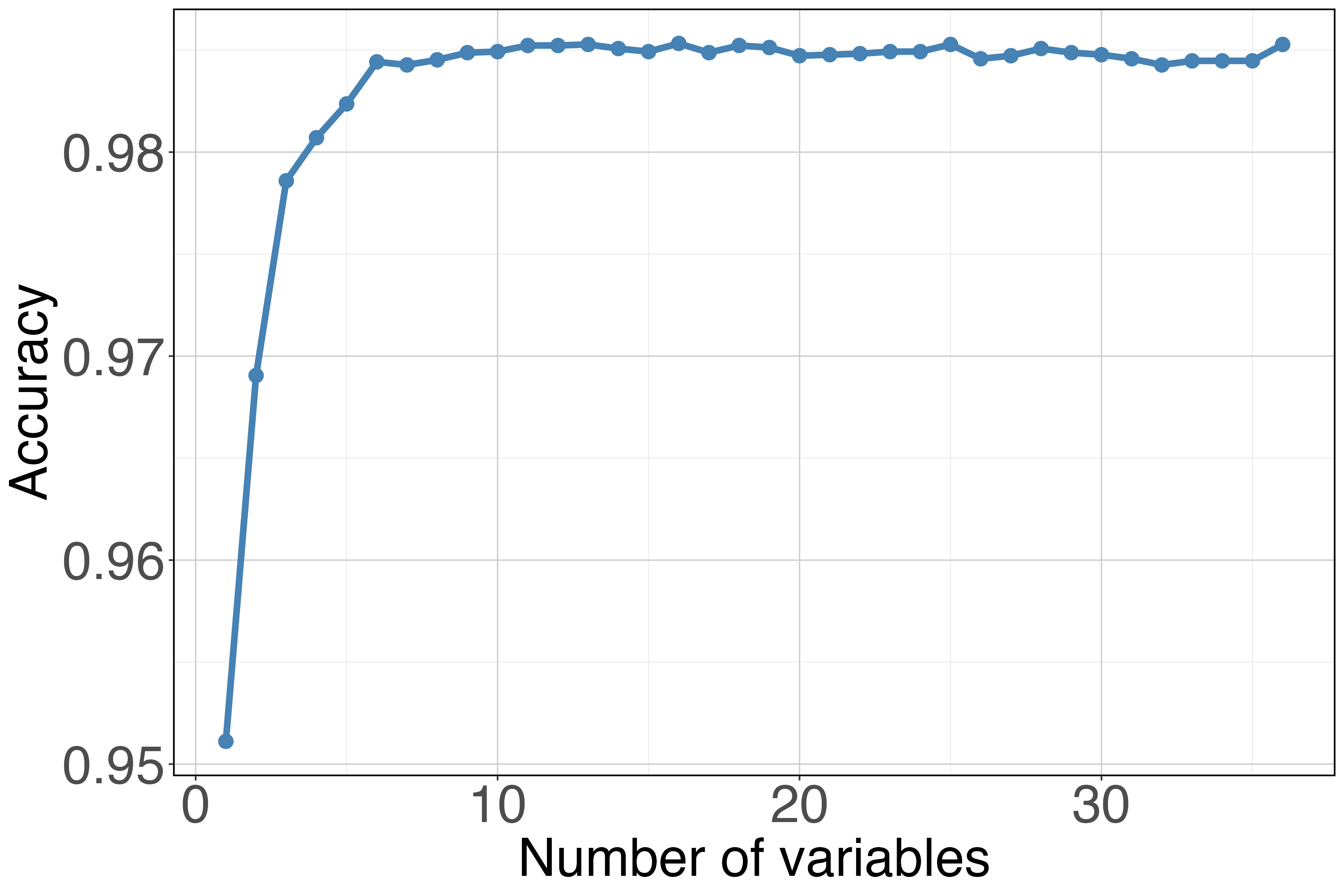} &
\includegraphics[width=0.49\textwidth]{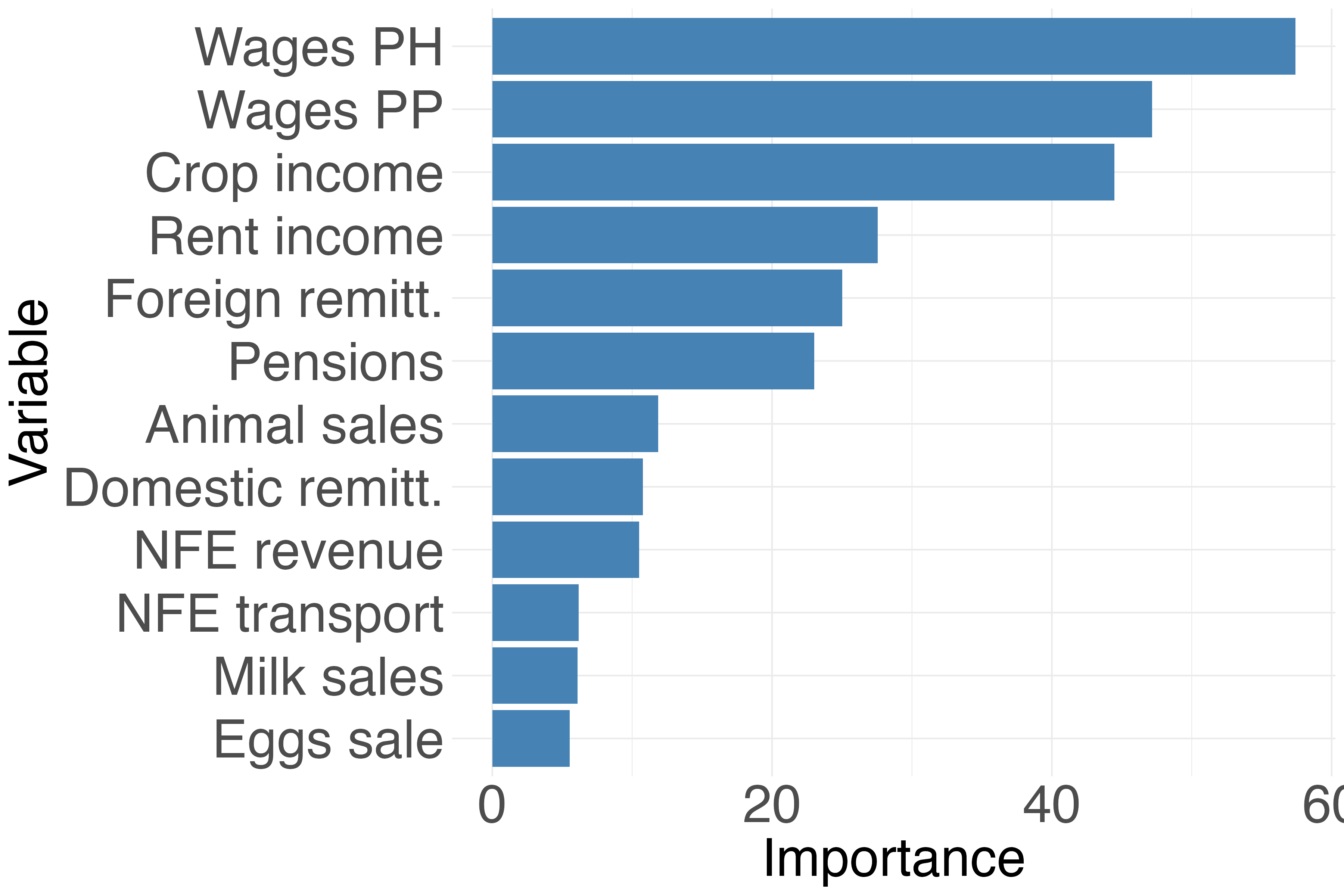} \\
\end{tabular}
\caption{Model accuracy and most relevant variables in predicting individuals' income position relative to the relative inequality line}
\label{fig:income_RIL}
\begin{flushleft}
\footnotesize \textit{Source:} Authors' compilation.
\end{flushleft}
\end{figure}

Unlike consumption, which is conventionally evaluated across both the post-harvest and post-planting periods to capture seasonal variation, earnings from the primary occupation represent the only income category available for both periods. Consequently, we include these two seasonal earning components as separate potential predictors.\footnote{We also estimated the model using a single aggregated income variable constructed as the average of post-planting and post-harvest earnings. The relationship between model accuracy and the number of predictors remains qualitatively similar to the results presented in Figures \ref{fig:income_quintile}-\ref{fig:income_RIL}.}

We first focus on predicting an individual's classification within the quintiles of the income distribution. The left panel of Figure \ref{fig:income_quintile} illustrates model accuracy as a function of the number of variables included. The results suggest that the maximum predictive accuracy for income quintiles converges around 80 percent. Notably, a model with just six variables surpasses the 75 percent accuracy threshold, whereas reaching the 80 percent plateau requires a 12-variable specification.

The right panel of Figure \ref{fig:income_quintile} displays the most influential predictors of income quintiles. The nine most important predictors are positive income streams derived from crop sales, wages, remittances, non-farm enterprise (NFE) revenue, livestock sales, pensions, and rental income. The subsequent categories consist of various production costs, specifically three related to NFEs and two associated with pesticide and fertilizer acquisitions.

The prominence of crop sales, wage earnings, remittances and non-farm enterprise revenues is consistent with the literature on rural livelihoods in Sub-Saharan Africa, which emphasises that household welfare is shaped by diversified income portfolios rather than by agricultural production alone \citep{BarrettReardonWebb2001}. The results therefore suggest that a reduced income module should preserve the main revenue-generating activities across farm, labour-market, enterprise and transfer channels.

We now turn our attention to the prediction of poverty status. Because the international poverty line is conventionally applied to welfare aggregates expressed in comparable purchasing-power-parity terms, we restrict the income analysis to a relative poverty threshold defined as 60 percent of the median income aggregate. As we focus here exclusively on income, we present only the results for the relative income poverty line, defined as 60 percent of the median income aggregate. The left panel of Figure \ref{fig:income_NLP} illustrates the predictive accuracy of our models as a function of the number of included variables. The results demonstrate that a remarkably parsimonious model can capture poverty status with high precision. Relying on only three predictors, the model correctly classifies more than 80 percent of individuals’ poverty status. When expanding the specification to just five variables, the predictive precision reaches 90 percent.

This high accuracy is consistent with the logic of supervised poverty-targeting tools, which exploit a small number of highly informative welfare correlates to classify households around a threshold \citep{CoadyGroshHoddinott2004}. However, the present ML approach differs from standard proxy means testing because the same reduced income variables are also evaluated for their ability to preserve broader distributional rankings.

An analysis of the feature importance rankings (right panel of Figure \ref{fig:income_NLP}) reveals striking structural consistency: the nine most influential variables for predicting poverty status coincide exactly in order of importance with those selected by the quintile models. This overlap carries important policy and methodological implications. First, from a data-collection perspective, it suggests that the core drivers of structural economic positioning remain identical whether modelling broad distribution brackets (quintiles) or a binary welfare threshold (poverty status). Consequently, the same reduced questionnaire could achieve dual analytical goals: mapping overall inequality and targeting the poor.\footnote{Crucially, this structural alignment may reflect the specific socioeconomic profile of Nigeria, and further cross-country validation is required before general scaling conclusions can be drawn.} Second, because these top nine predictors are predominantly positive income streams (such as crop sales, wages, and non-farm enterprise revenues) rather than agricultural input or production costs, it implies that poverty status in this context is fundamentally driven by severe revenue deficits rather than variations in operational expenses. This finding suggests that social protection programmes and targeted interventions should prioritise income-generating capacity over input-subsidisation schemes alone.

We conclude this section by analysing the predictive accuracy of our models in identifying whether individuals fall above or below the relative inequality line (RIL) corresponding to the Gini index. As illustrated in Figure \ref{fig:income_RIL}, and consistent with our previous findings regarding absolute poverty lines, high levels of accuracy are achieved using a very limited set of variables. Notably, post-harvest earnings alone successfully predict an individual’s position relative to this threshold in 94 percent of cases. Incorporating post-planting earnings into the specification further increases the predictive precision to over 96 percent.

While several of the most influential predictors coincide with those observed in the quintile and national poverty line models, their relative rankings vary substantially. For this threshold, labour earnings eclipse crop sales in predictive power. Furthermore, pensions and rental income assume a more prominent position, displacing remittances and livestock sales relative to their rankings in previous models.

\section{Robustness Checks}
\label{sec:Robust}

To ensure that the predictive performance of our parsimonious models is not an artefact of specific demographic assumptions, it is crucial to evaluate their robustness across alternative definitions of individual welfare. The baseline assumption of per capita income implies a completely linear relationship between household size and resource requirements. However, in settings with significant shared resources (such as housing, fuel, or durable goods) or variations in demographic composition (such as the ratio of children to adults), a per capita metric may underestimate individual welfare in larger households. Conversely, non-linear equivalence scales acknowledge economies of scale within the household unit. Because our machine learning models rely heavily on variables deeply intertwined with household size (such as total crop sales, labour earnings, and specific production costs) shifting the target variable’s distribution via different scales could theoretically alter total predictive accuracy.

\begin{figure}[htbp]
\centering
    \begin{subfigure}[b]{0.49\textwidth}
        \centering
        \includegraphics[width=\textwidth]{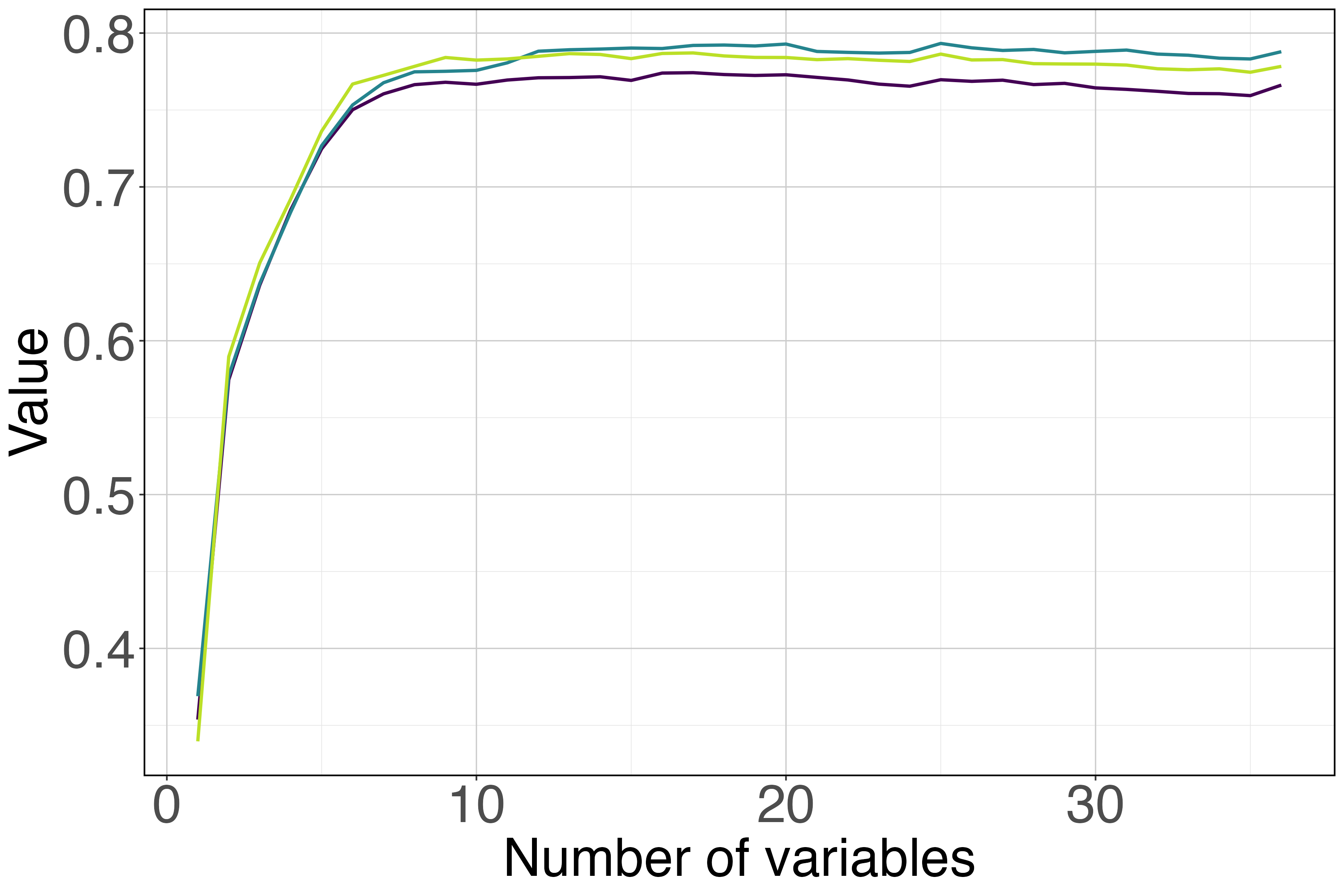}
        \caption{Quintiles}
        \label{fig:ril_accuracy}
    \end{subfigure}
    \hfill
    \begin{subfigure}[b]{0.49\textwidth}
        \centering
        \includegraphics[width=\textwidth]{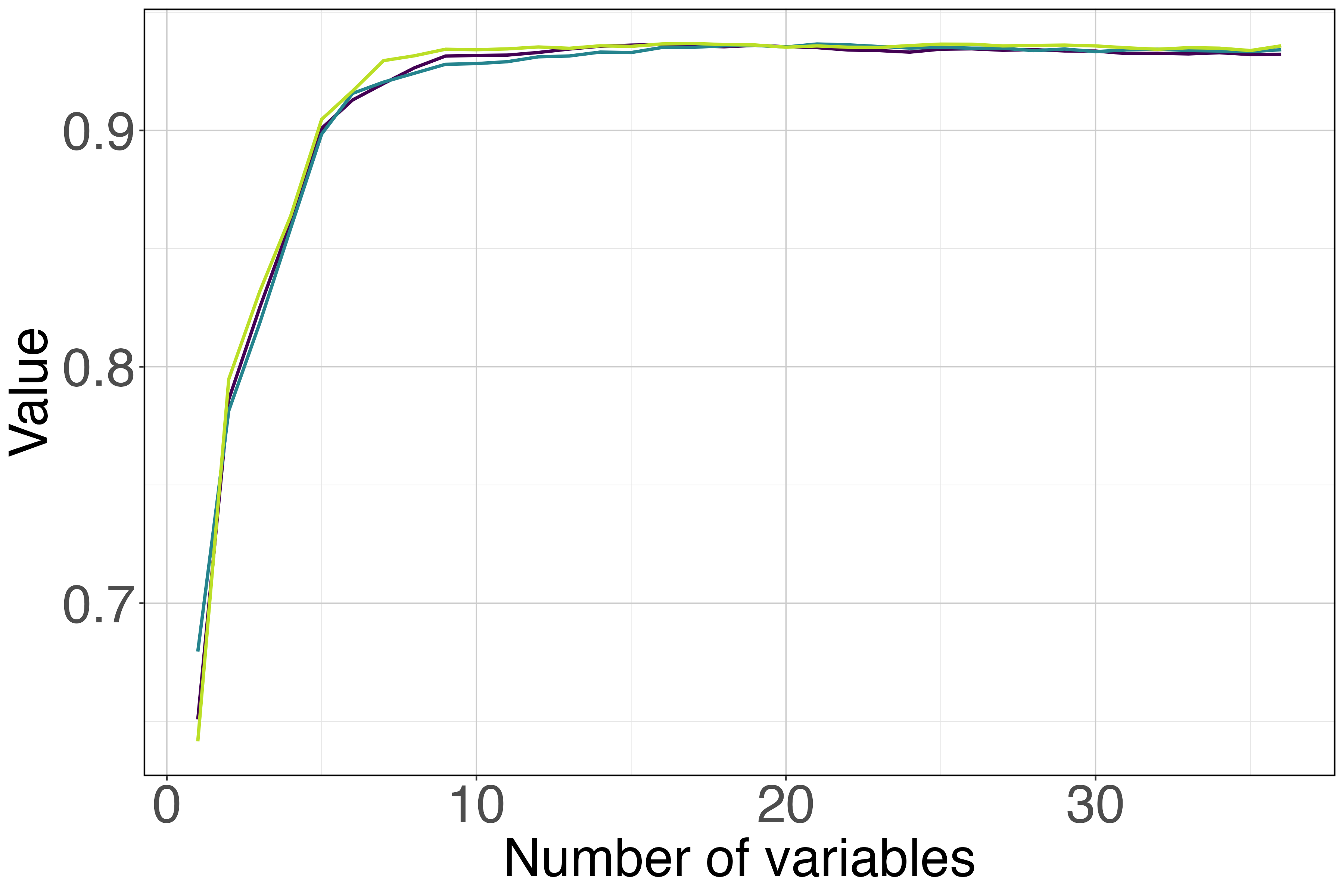}
        \caption{National poverty line}
        \label{fig:ril_variables}
    \end{subfigure}
    
    \vspace{0.5cm} 
    
    \begin{subfigure}[b]{0.70\textwidth} 
        \centering
        \includegraphics[width=\textwidth]{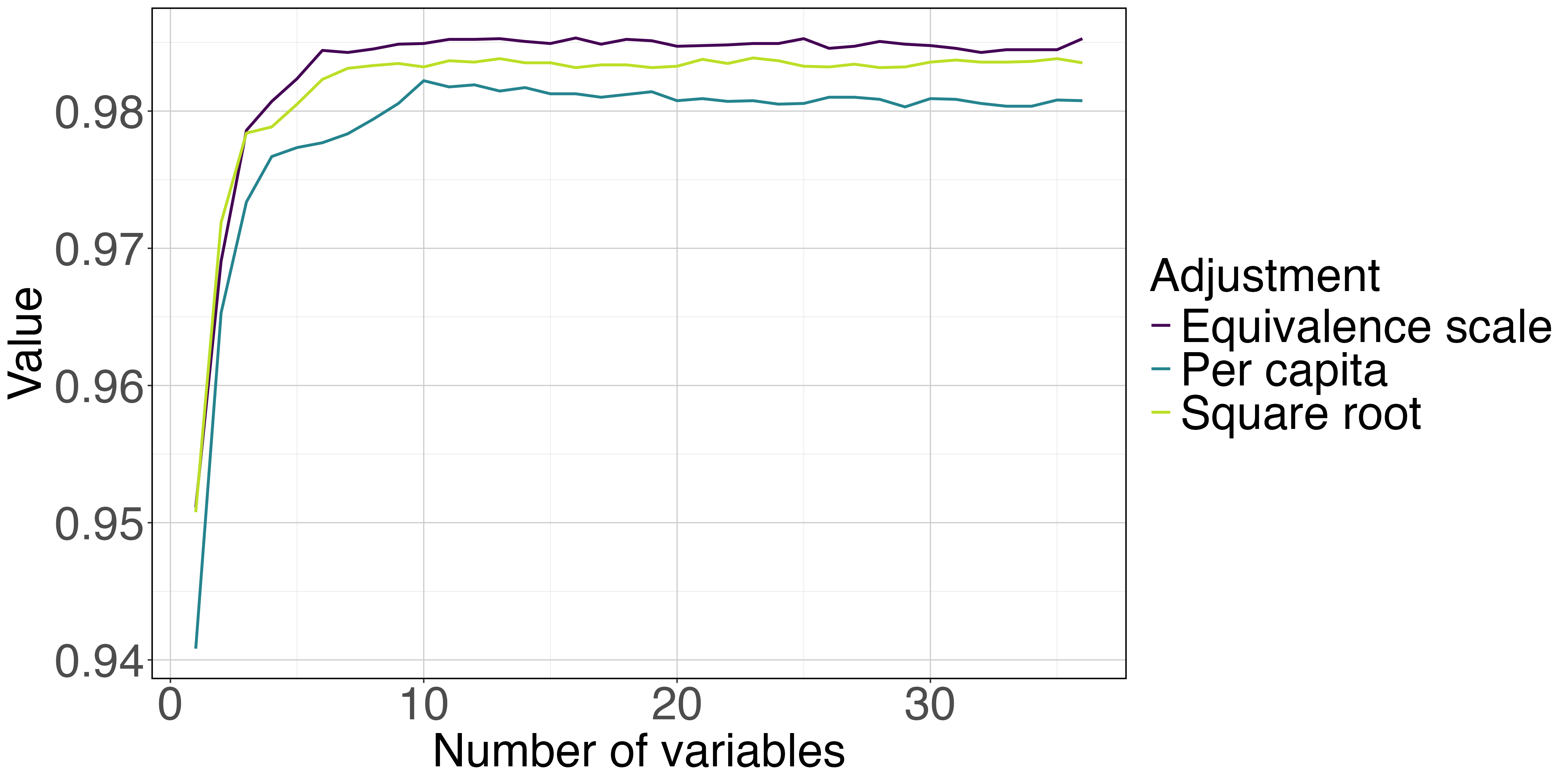} 
        \caption{Relative inequality line}
        \label{fig:ril_bottom}
    \end{subfigure}

\caption{Model accuracy in predicting individuals' income position for different equivalence scales}
\label{fig:roboust_eq_scale}
\begin{flushleft}
\footnotesize \textit{Source:} Authors' compilation.
\end{flushleft}
\end{figure}

Figure \ref{fig:roboust_eq_scale} illustrates the predictive accuracy of the models as a function of the number of included variables for the three adjustment methods: per capita, square root, and the OECD equivalence scale. For the three alternative classifications by income groups (income quintiles in Panel a, the national poverty line in Panel b, and the relative inequality line in Panel c) the models exhibit remarkable structural stability. The learning curves closely mirror one another, demonstrating that the speed of model convergence and the ultimate accuracy peak are virtually invariant to the chosen equivalence scale.

The maximum accuracy for predicting income quintiles consistently stabilises at approximately 80 percent across all scales, with a model of just six variables comfortably passing the 75 percent accuracy threshold in every scenario. This consistency is even more pronounced when analysing the national poverty line. The models achieve a matching 90 percent accuracy with only five variables.

Relative inequality line models displayed in Panel (c) show a marginal divergence, though the overall precision remains exceptionally high. Here, the equivalence-scale specifications yield slightly superior predictive performance compared to the baseline per capita metric, maximising accuracy at roughly 98.5 percent. The per capita line trails marginally behind, stabilizing just above 98 percent after the inclusion of ten variables. 

\begin{figure}[htbp]
\centering
\begin{tabular}{ll}
(a) Accuracy & (b) Variables \\
\includegraphics[width=0.49\textwidth]{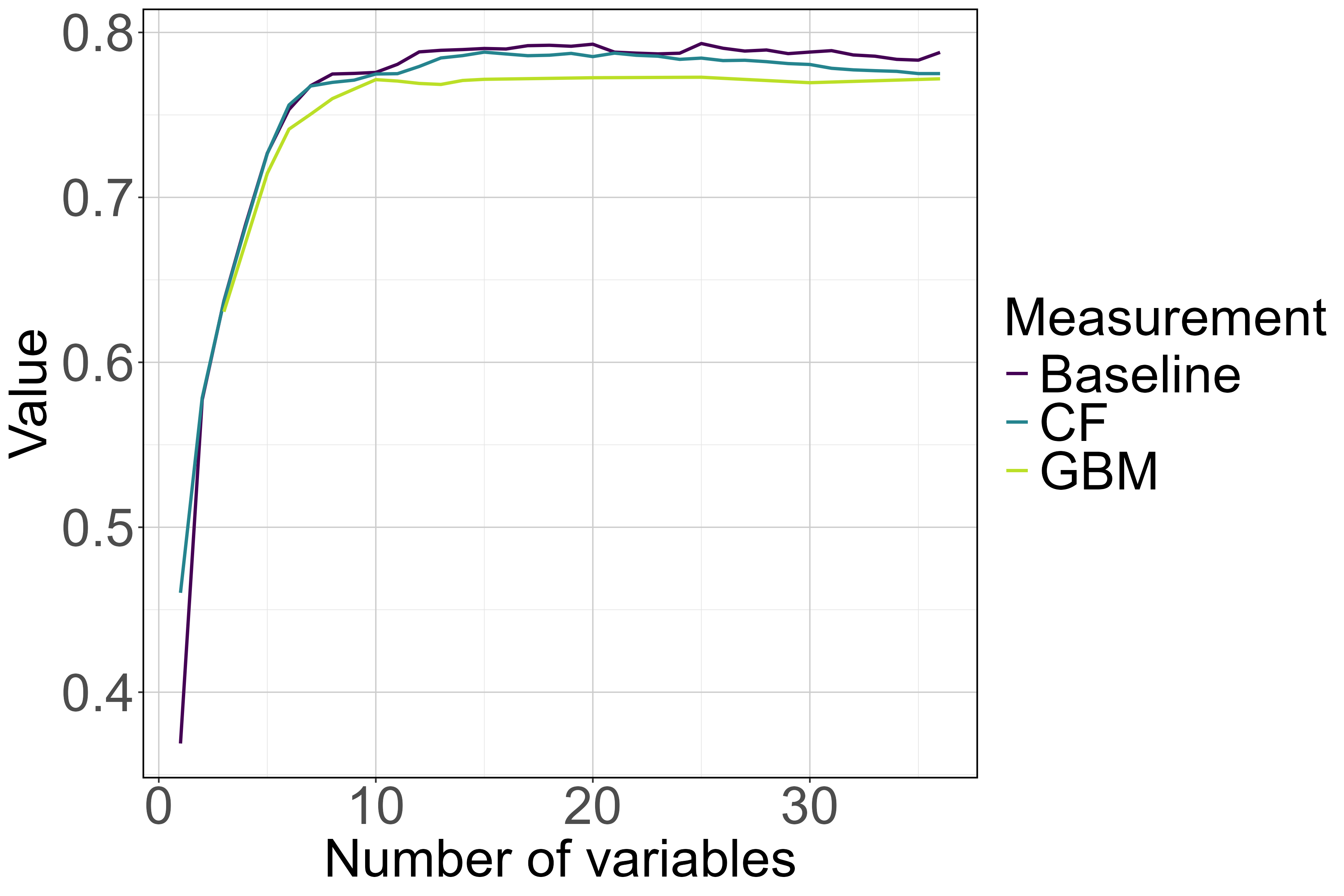} &
\includegraphics[width=0.49\textwidth]{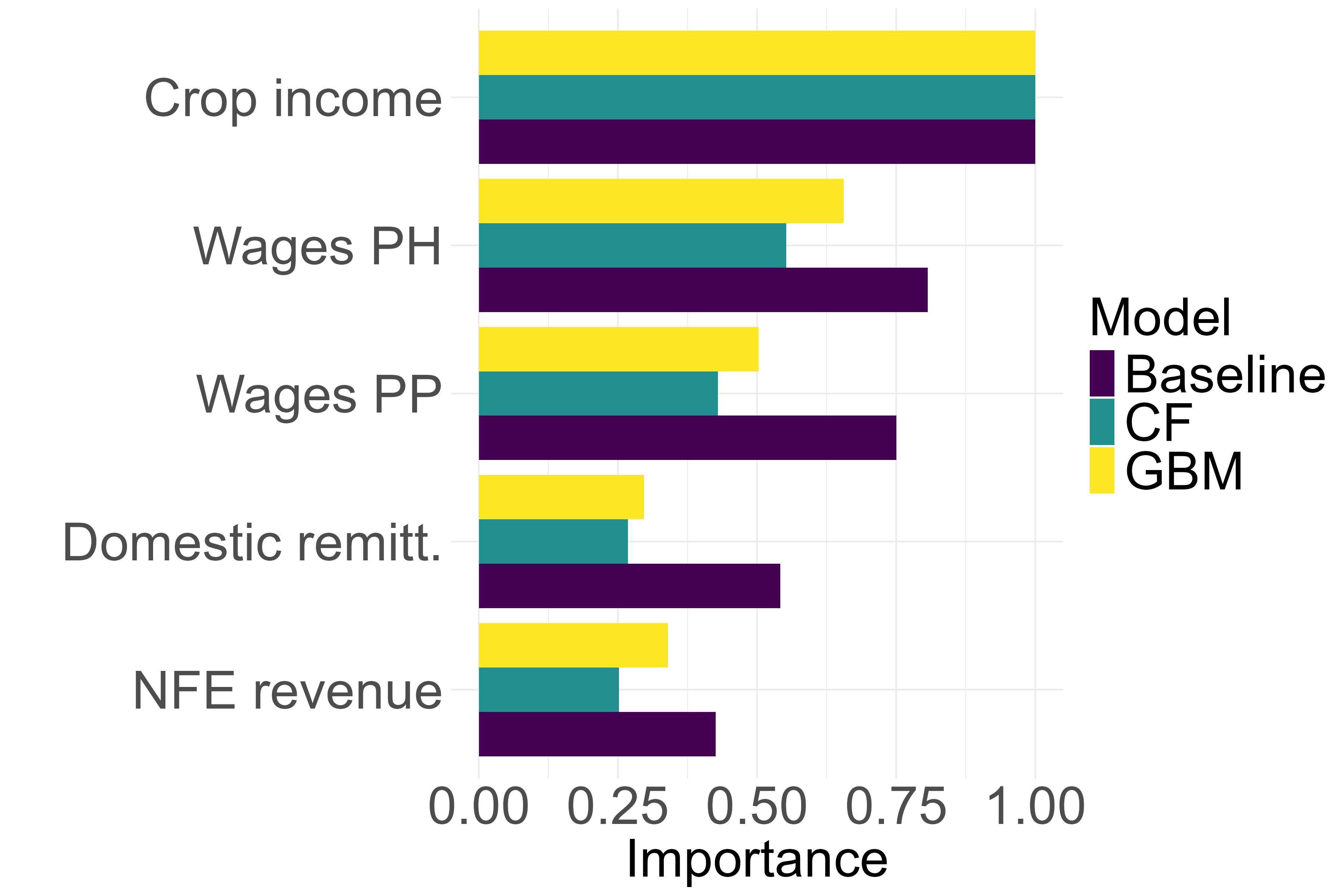} \\
\end{tabular}
\caption{Model accuracy and most relevant variables in predicting income quintile classification across alternative machine-learning methods}
\label{fig:alt_methods}

\begin{flushleft}
\footnotesize 
\textit{Source:} Authors' compilation. \\
\textit{Note:} Variable importance measures were normalised so that 1 represents the maximum importance, ensuring comparability across methods.
\end{flushleft}
\end{figure}

For robustness purposes, alternative machine-learning algorithms are considered in addition to the baseline random forest specification. The objective is to assess whether the predictive performance and the ranking of the most relevant variables are sensitive to the choice of classification algorithm. The baseline specification uses a conventional random forest classifier. Random forests construct a large number of decision trees using bootstrap samples and random subsets of predictors and aggregate predictions across trees. This approach has several advantages, including high predictive accuracy, robustness to overfitting, the ability to capture complex nonlinear relationships and interactions, and relatively limited tuning requirements. However, conventional random forests may exhibit some limitations. Variable importance measures can be biased toward variables with many possible splitting values or variables displaying high variance. Furthermore, standard random forests treat predictors symmetrically and do not explicitly account for potential biases arising from correlated variables.

To address these limitations, two alternative models are considered. Conditional inference forests replace the split-selection procedure used in conventional random forests with a statistical testing framework \citep{HothornHornikZeileis2006}. Instead of selecting variables according to impurity reduction criteria, splits are chosen based on permutation tests. This procedure reduces the tendency of standard random forests to favour predictors with many possible split points or larger variability \citep{StroblEtAl2007}. Consequently, conditional inference forests may produce more reliable variable importance rankings when predictors differ substantially in scale or distribution. However, these models are computationally more demanding and may occasionally yield slightly lower predictive performance than conventional random forests.

Gradient boosting (GB) constructs trees sequentially rather than independently. Each new tree is estimated to improve the prediction errors generated by previous trees \citep{Friedman2001}. This iterative process often leads to stronger predictive performance and may capture subtle non-linear patterns that standard random forests fail to identify. Nevertheless, GB models require more extensive hyperparameter tuning and may be more sensitive to overfitting if the number of trees, learning rate, or tree depth are not appropriately selected.

Figure~\ref{fig:alt_methods} illustrates the predictive accuracy of the alternative estimation frameworks alongside our baseline methodology in predicting individuals’ positions within the income distribution. The results indicate that the ensemble methods based on random forests exhibit comparable classification performance, while gradient boosting yields slightly lower performance. 

The right panel of Figure~\ref{fig:alt_methods} presents the relative importance scores of the top five predictors driving these income quintile classifications. Because raw feature importance figures are not directly comparable across different methodologies, we normalised these, so that the maximum value equals one within each method. Interestingly, all three approaches identify the same key predictors of income quintile classification. Crop income emerges as the most important variable, followed by earnings in the post-harvest period and, closely thereafter, earnings in the post-planting period. For the forest-based methodologies, the next most critical predictors are domestic remittances within Nigeria, followed by the revenue generated from non-farm enterprises. In the case of gradient boosting, however, the relative importance of these last two variables is reversed.

\section{Conclusion}
\label{sec:Conclusion}

Reliable poverty and inequality measurement requires detailed information on household income and consumption, yet collecting such data through conventional household income and expenditure surveys is costly, time-consuming and demanding for both statistical agencies and respondents. This paper has examined whether reduced survey instruments can retain sufficient distributional information to support poverty and inequality measurement and monitoring. Using data from the Nigeria General Household Survey, we applied Random Forest Recursive Feature Elimination to identify a parsimonious set of income sources and consumption categories capable of reproducing key classifications derived from full welfare aggregates.

The analysis focused on three distributional outcomes: poverty status, quintile classification and position relative to the Gini-based inequality line. The results show that reduced models can achieve strong predictive performance, although accuracy varies across welfare concepts and information sets. For consumption, poverty status and inequality-line position are classified accurately with a small number of expenditure categories. Quintile classification is more demanding, especially when annual consumption is predicted from a single seasonal visit, but the aggregated consumption specification still achieves high accuracy for seasonal welfare rankings. The results also show that broader consumption categories can outperform more granular item-level information, suggesting that aggregation may reduce idiosyncratic reporting noise while preserving the budget components most relevant for welfare classification.

The consumption results highlight the central role of food expenditure and housing-related costs in reduced-data welfare measurement. Food, meals consumed outside the home and rent are consistently among the most informative predictors of consumption quintiles, while roots, vegetables and grains are especially important for poverty classification. The analysis also shows that seasonal information matters. Post-planting and post-harvest models differ in the relative importance assigned to essential food items, discretionary spending and housing-related expenditure, reinforcing the need to account for agricultural seasonality when designing shorter welfare modules in rural and semi-rural contexts.

The income results point to a similarly concentrated set of informative variables. Crop sales, labour earnings, remittances, non-farm enterprise revenues, livestock sales, pensions and rental income account for much of the predictive information needed to classify individuals within the income distribution. Income poverty status can be predicted with high accuracy using only a few variables, while position relative to the Gini-based inequality line is captured particularly well by labour earnings. These findings suggest that reduced income modules should preserve information on the main revenue-generating activities across agriculture, labour markets, household enterprises and transfers, rather than focusing narrowly on a single source of income.

The paper contributes to the literature on welfare measurement in two ways. First, it moves beyond approaches designed primarily for poverty targeting or relative wealth ranking by evaluating whether a reduced set of variables can preserve multiple distributional classifications simultaneously. Second, it provides evidence that machine-learning methods can support survey design by identifying the smallest set of questions that achieves a high and stable level of predictive performance. This is particularly relevant for countries where the cost of full household income and expenditure surveys limits the frequency of poverty and inequality monitoring.

Several caveats are important. The analysis is based on a single country, and the selected predictors may partly reflect Nigeria's specific economic structure, agricultural calendar and survey design. The results should therefore be interpreted as evidence of feasibility rather than as a universal reduced questionnaire. Cross-country validation is needed to assess whether similar sets of variables perform well in other low- and middle-income contexts. In addition, the current exercise evaluates classification accuracy rather than the direct recovery of continuous welfare levels or inequality indices. Reduced instruments may classify households accurately around relevant thresholds while still compressing the tails of the distribution, an issue that deserves further investigation.

Overall, the findings suggest that carefully designed reduced survey instruments, supported by supervised feature selection methods, can retain a substantial amount of the information required for poverty and inequality monitoring. Such tools should not replace full household surveys, which remain essential for constructing benchmark welfare aggregates and recalibrating reduced instruments. However, they can complement comprehensive surveys by enabling more frequent, lower-cost and operationally feasible monitoring of welfare classifications between full survey rounds. In this sense, RF-RFE and related machine-learning methods offer a promising route for strengthening statistical capacity and improving the timeliness of distributional evidence in resource-constrained settings.

\section*{Acknowledgements}
The authors are grateful to participants of the WIDER Development Conference 2026 "Green Industrialisation and Inclusive Growth in a Fractured World Order" in New Delhi, India for their comments and suggestions on early versions of this paper. VJ acknowledges financial support from the I+D+i project Ref. PID2024-156871NB-I00 financed by MICIU/AEI/10.13039/501100011033/FEDER,UE. All errors are solely the responsibility of the authors.

\clearpage
\bibliography{references}

@article{Breiman2001,
	author = {Breiman, Leo},
	doi = {10.1023/A:1010933404324},
	journal = {Machine Learning},
	number = {1},
	pages = {5--32},
	title = {Random Forests},
	volume = {45},
	year = {2001}}

@techreport{LopezNoval2024,
	address = {Helsinki},
	author = {Lopez-Noval, B. and Ni{\~n}o-Zaraz{\'u}a, M. and Roope, L. and Tarp, F.},
	doi = {10.35188/UNU-WIDER/2024/540-0},
	institution = {UNU-WIDER},
	number = {2024/77},
	title = {From the bottom 40 to inequality lines: Sharing prosperity globally and domestically},
	type = {WIDER Working Paper},
	url = {https://doi.org/10.35188/UNU-WIDER/2024/540-0},
	year = {2024}}

@techreport{nbs2020,
	address = {Abuja},
	author = {{National Bureau of Statistics}},
	title = {2019 Poverty and Inequality in Nigeria: Executive Summary},
	year = {2020}}

@techreport{ravallion1998,
	address = {Washington, DC},
	author = {Ravallion, M.},
	institution = {World Bank},
	number = {133},
	title = {Poverty Lines in Theory and Practice},
	type = {Living Standards Measurement Study Working Paper},
	year = {1998}}

@book{worldbank2022,
	address = {Washington, DC},
	author = {{World Bank}},
	publisher = {World Bank},
	title = {Poverty and Shared Prosperity 2022: Correcting Course},
	year = {2022}}

@book{DeatonZaidi2002,
	address = {Washington, DC},
	author = {Deaton, A. and Zaidi, S.},
	publisher = {World Bank},
	series = {LSMS Working Paper No. 135},
	title = {Guidelines for Constructing Consumption Aggregates for Welfare Analysis},
	year = {2002}}

@book{Friedman2001,
	address = {New York},
	author = {Friedman, J. and Hastie, T. and Tibshirani, R.},
	publisher = {Springer},
	title = {The Elements of Statistical Learning},
	year = {2001}}

@inproceedings{Guyon2002,
	author = {Guyon, I. and Weston, J. and Barnhill, S. and Vapnik, V.},
	booktitle = {Machine Learning},
	number = {1-3},
	pages = {389--422},
	title = {Gene Selection for Cancer Classification using Support Vector Machines},
	volume = {46},
	year = {2002}}

@article{Lanjouw1995,
	author = {Lanjouw, P. and Ravallion, M.},
	journal = {The Economic Journal},
	number = {433},
	pages = {1415--1434},
	title = {Poverty and Household Size},
	volume = {105},
	year = {1995}}

@book{beegle2012methods,
	author = {Beegle, Kathleen and De Weerdt, Joachim and Friedman, Jed and Gibson, John},
	publisher = {The World Bank},
	title = {Methods of household consumption measurement through surveys in developing countries},
	year = {2012}}

@book{deaton1997analysis,
	author = {Deaton, Angus},
	publisher = {World Bank Publications},
	title = {The analysis of household surveys: a microeconometric approach to development policy},
	year = {1997}}

@book{deaton1980economics,
	author = {Deaton, Angus and Muellbauer, John},
	publisher = {Cambridge university press},
	title = {Economics and consumer behavior},
	year = {1980}}

@article{hagenaars1994statistical,
	author = {Hagenaars, Aldi and et al.},
	journal = {Journal of Official Statistics},
	number = {1},
	pages = {53--64},
	title = {Statistical matching of microdata files: Theory and practice},
	volume = {10},
	year = {1994}}

@book{canberragroup2011,
	author = {{Canberra Group}},
	publisher = {United Nations Economic Commission for Europe},
	title = {Canberra Group Handbook on Household Income Statistics},
	year = {2011}}

@article{atkinson1970measurement,
	author = {Atkinson, Anthony B.},
	journal = {Journal of Economic Theory},
	number = {3},
	pages = {244--263},
	title = {On the Measurement of Inequality},
	volume = {2},
	year = {1970}}

@book{piketty2014capital,
	author = {Piketty, Thomas},
	publisher = {Harvard University Press},
	title = {Capital in the Twenty-First Century},
	year = {2014}}

@book{cowell2011measuring,
	author = {Cowell, Frank A.},
	edition = {3},
	publisher = {Oxford University Press},
	title = {Measuring Inequality},
	year = {2011}}

@book{bourguignon2015global,
	author = {Bourguignon, Fran{\c{c}}ois},
	publisher = {Princeton University Press},
	title = {The Globalization of Inequality},
	year = {2015}}

@book{deaton2002guidelines,
	author = {Deaton, Angus and Zaidi, Salman},
	publisher = {World Bank},
	title = {Guidelines for Constructing Consumption Aggregates for Welfare Analysis},
	year = {2002}}

@book{ravallion1998poverty,
	author = {Ravallion, Martin},
	publisher = {World Bank},
	title = {Poverty Lines in Theory and Practice},
	year = {1998}}

@techreport{worldbank2023poverty,
	address = {Washington, DC},
	author = {{Independent Evaluation Group}},
	institution = {World Bank},
	title = {Poverty Mapping: Innovative Approaches to Creating Poverty Maps from New Data Sources},
	year = {2023}}

@article{grosh1998data,
	author = {Grosh, Margaret and Glewwe, Paul},
	journal = {Journal of Economic Perspectives},
	number = {1},
	pages = {187--196},
	title = {Data Watch: The World Bank's Living Standards Measurement Study Household Surveys},
	volume = {12},
	year = {1998}}

@techreport{kilic2017costing,
	author = {Kilic, Talip and Serajuddin, Umar and Uematsu, Hiroki and Yoshida, Nobuo},
	institution = {World Bank},
	number = {7951},
	title = {Costing Household Surveys for Monitoring Progress Toward Ending Extreme Poverty and Boosting Shared Prosperity},
	type = {Policy Research Working Paper},
	year = {2017}}

@book{Hastie2009,
	author = {Hastie, Trevor and Tibshirani, Robert and Friedman, Jerome},
	publisher = {Springer},
	title = {The Elements of Statistical Learning},
	year = {2009}}

@book{Deaton1997,
	address = {Baltimore},
	author = {Deaton, Angus},
	publisher = {Johns Hopkins University Press for the World Bank},
	title = {The Analysis of Household Surveys: A Microeconometric Approach to Development Policy},
	year = {1997}}

@article{FilmerPritchett2001,
	author = {Filmer, Deon and Pritchett, Lant H.},
	journal = {World Bank Economic Review},
	number = {1},
	pages = {115--132},
	title = {Estimating Wealth Effects without Expenditure Data---or Tears: An Application to Educational Enrollments in States of India},
	volume = {15},
	year = {2001}}

@article{SahnStifel2003,
	author = {Sahn, David E. and Stifel, David},
	journal = {Review of Income and Wealth},
	number = {4},
	pages = {463--489},
	title = {Exploring Alternative Measures of Welfare in the Absence of Expenditure Data},
	volume = {49},
	year = {2003}}

@techreport{GroshBaker1995,
	author = {Grosh, Margaret and Baker, Judy},
	institution = {World Bank, Living Standards Measurement Study (LSMS)},
	number = {118},
	title = {Proxy Means Tests for Targeting Social Programs: Simulations and Speculation},
	year = {1995}}

@article{Beegle2012,
	author = {Beegle, Kathleen and De Weerdt, Joachim and Friedman, Jed and Gibson, John},
	journal = {Journal of Development Economics},
	number = {1},
	pages = {3--18},
	title = {Methods of Household Consumption Measurement through Surveys: Experimental Results from Tanzania},
	volume = {98},
	year = {2012}}

@article{GibsonKim2007,
	author = {Gibson, John and Kim, Bonggeun},
	journal = {American Journal of Agricultural Economics},
	number = {2},
	pages = {473--489},
	title = {Measurement Error in Recall Surveys and the Relationship between Household Size and Food Demand},
	volume = {89},
	year = {2007}}

@article{Moore2000,
	author = {Moore, Jeffrey C. and Stinson, Linda L. and Welniak, Edward J.},
	journal = {Journal of Official Statistics},
	number = {4},
	pages = {331--361},
	title = {Income Measurement Error in Surveys: A Review},
	volume = {16},
	year = {2000}}

@incollection{Bound2001,
	author = {Bound, John and Brown, Charles and Mathiowetz, Nancy},
	booktitle = {Handbook of Econometrics, Volume 5},
	editor = {Heckman, James J. and Leamer, Edward},
	pages = {3705--3843},
	publisher = {Elsevier},
	title = {Measurement Error in Survey Data},
	year = {2001}}

@article{PradhanRavallion2000,
	author = {Pradhan, Menno and Ravallion, Martin},
	journal = {Review of Economics and Statistics},
	number = {3},
	pages = {462--471},
	title = {Measuring Poverty Using Qualitative Perceptions of Consumption Adequacy},
	volume = {82},
	year = {2000}}

@article{RavallionLokshin2002,
	author = {Ravallion, Martin and Lokshin, Michael},
	journal = {European Economic Review},
	number = {8},
	pages = {1453--1473},
	title = {Self-Rated Economic Welfare in Russia},
	volume = {46},
	year = {2002}}

@article{montgomery2000measuring,
	author = {Montgomery, Mark R. and Gragnolati, Michele and Burke, Kathleen A. and Paredes, Edmundo},
	journal = {Demography},
	number = {2},
	pages = {155--174},
	title = {Measuring Living Standards with Proxy Variables},
	volume = {37},
	year = {2000}}

@article{mckenzie2005measuring,
	author = {McKenzie, David J.},
	journal = {Journal of Population Economics},
	number = {2},
	pages = {229--260},
	title = {Measuring Inequality with Asset Indicators},
	volume = {18},
	year = {2005}}

@article{vyas2006constructing,
	author = {Vyas, Seema and Kumaranayake, Lilani},
	journal = {Health Policy and Planning},
	number = {6},
	pages = {459--468},
	title = {Constructing Socio-Economic Status Indices: How to Use Principal Components Analysis},
	volume = {21},
	year = {2006}}

@article{chakraborty2016equitytool,
	author = {Chakraborty, Nirali M. and Fry, Kathryn and Behl, Ritu and Longfield, Kim},
	journal = {Global Health: Science and Practice},
	number = {4},
	pages = {503--515},
	title = {A New Approach to Measure Socioeconomic Status in Low- and Middle-Income Countries Using the EquityTool},
	volume = {4},
	year = {2016}}

@article{howe2008measuring,
	author = {Howe, Laura D. and Hargreaves, James R. and Huttly, Sharon R. A.},
	journal = {International Journal of Epidemiology},
	number = {4},
	pages = {871--881},
	title = {Measuring Socio-Economic Position for Epidemiological Studies in Low- and Middle-Income Countries: A Methods of Measurement in Epidemiology Paper},
	volume = {37},
	year = {2008}}

@article{harttgen2013comparison,
	author = {Harttgen, Kenneth and Vollmer, Sebastian},
	journal = {World Development},
	pages = {1--13},
	title = {A Comparison of Asset Indices and Consumption Expenditure as Welfare Measures in Developing Countries},
	volume = {41},
	year = {2013}}

@article{smits2011towards,
	author = {Smits, Jeroen and Steendijk, Roel},
	journal = {Social Science \& Medicine},
	number = {3},
	pages = {353--361},
	title = {Towards a Better Understanding of Socioeconomic Inequality in Child Health: Asset-Based Wealth Indices Versus Consumption-Based Measures},
	volume = {72},
	year = {2011}}

@book{Schreiner2010,
	address = {St. Louis, MO},
	author = {Schreiner, Mark},
	publisher = {Microfinance Risk Management, L.L.C.},
	title = {A Simple Poverty Scorecard for {Country}: Design and Use},
	year = {2010}}

@article{Hurd2004,
	author = {Hurd, Michael D. and Kreuter, Frauke and Winter, Joachim},
	journal = {Journal of Official Statistics},
	number = {3},
	pages = {595--612},
	title = {Unfolding Brackets in Surveys: Efficiency Gains and Reduced Nonresponse},
	volume = {20},
	year = {2004}}

@techreport{Kennickell1998,
	author = {Kennickell, Arthur B.},
	institution = {Board of Governors of the Federal Reserve System},
	note = {Available at: https://www.federalreserve.gov/econres/scfindex.htm},
	title = {Multiple Imputation in the Survey of Consumer Finances},
	year = {1998}}

@article{CowellMehta1982,
	author = {Cowell, Frank A. and Mehta, F.},
	journal = {Review of Economic Studies},
	number = {2},
	pages = {273--290},
	title = {The Estimation and Interpolation of Inequality Measures},
	volume = {49},
	year = {1982}}

@article{CowellVictoriaFeser1996,
	author = {Cowell, Frank A. and Victoria-Feser, Maria-Pia},
	journal = {Econometrica},
	number = {1},
	pages = {77--101},
	title = {Robustness Properties of Inequality Measures},
	volume = {64},
	year = {1996}}

@book{Ravallion1994,
	address = {Chur, Switzerland},
	author = {Ravallion, Martin},
	publisher = {Harwood Academic Publishers},
	title = {Poverty Comparisons},
	year = {1994}}

@misc{NBSGHSPanel2019,
	author = {{National Bureau of Statistics}},
	howpublished = {World Bank Microdata Library},
	note = {Reference ID: NGA\_2018\_GHSP-W4\_v03\_M},
	title = {{Nigeria General Household Survey-Panel 2018/19, Wave 4}},
	year = {2019}}

@book{GroshGlewwe2000,
	address = {Washington, DC},
	editor = {Grosh, Margaret and Glewwe, Paul},
	publisher = {World Bank},
	title = {Designing Household Survey Questionnaires for Developing Countries: Lessons from 15 Years of the Living Standards Measurement Study},
	year = {2000}}

@article{CarlettoEtAl2015,
	author = {Carletto, Calogero and Jolliffe, Dean and Banerjee, Raka},
	doi = {10.1080/00220388.2014.968140},
	journal = {The Journal of Development Studies},
	number = {2},
	pages = {133--148},
	title = {From Tragedy to Renaissance: Improving Agricultural Data for Better Policies},
	volume = {51},
	year = {2015}}

@article{KohaviJohn1997,
	author = {Kohavi, Ron and John, George H.},
	doi = {10.1016/S0004-3702(97)00043-X},
	journal = {Artificial Intelligence},
	number = {1--2},
	pages = {273--324},
	title = {Wrappers for Feature Subset Selection},
	volume = {97},
	year = {1997}}

@book{OECD2011,
	address = {Paris},
	author = {{OECD}},
	doi = {10.1787/soc_glance-2011-en},
	publisher = {OECD Publishing},
	title = {Society at a Glance 2011: OECD Social Indicators},
	year = {2011}}

@misc{Eurostat2021,
	author = {{Eurostat}},
	howpublished = {Statistics Explained},
	note = {At-risk-of-poverty threshold defined as 60 percent of national median equivalised disposable income after social transfers},
	title = {Glossary: At-risk-of-poverty rate},
	year = {2021}}

@article{BeegleEtAl2012,
	author = {Beegle, Kathleen and De Weerdt, Joachim and Friedman, Jed and Gibson, John},
	doi = {10.1016/j.jdeveco.2011.11.001},
	journal = {Journal of Development Economics},
	number = {1},
	pages = {3--18},
	title = {Methods of Household Consumption Measurement through Surveys: Experimental Results from Tanzania},
	volume = {98},
	year = {2012}}

@book{CoadyGroshHoddinott2004,
	address = {Washington, DC},
	author = {Coady, David and Grosh, Margaret and Hoddinott, John},
	publisher = {World Bank},
	title = {Targeting of Transfers in Developing Countries: Review of Lessons and Experience},
	year = {2004}}

@article{Paxson1993,
	author = {Paxson, Christina H.},
	doi = {10.1086/261865},
	journal = {Journal of Political Economy},
	number = {1},
	pages = {39--72},
	title = {Consumption and Income Seasonality in Thailand},
	volume = {101},
	year = {1993}}

@article{DerconKrishnan2000,
	author = {Dercon, Stefan and Krishnan, Pramila},
	doi = {10.1080/00220380008422653},
	journal = {Journal of Development Studies},
	number = {6},
	pages = {25--53},
	title = {Vulnerability, Seasonality and Poverty in Ethiopia},
	volume = {36},
	year = {2000}}

@article{BarrettReardonWebb2001,
	author = {Barrett, Christopher B. and Reardon, Thomas and Webb, Patrick},
	doi = {10.1016/S0306-9192(01)00014-8},
	journal = {Food Policy},
	number = {4},
	pages = {315--331},
	title = {Nonfarm Income Diversification and Household Livelihood Strategies in Rural Africa: Concepts, Dynamics, and Policy Implications},
	volume = {26},
	year = {2001}}

@article{Roope2021,
	author = {Roope, Laurence S. J.},
	doi = {10.1371/journal.pone.0248178},
	journal = {PLOS ONE},
	number = {3},
	pages = {e0248178},
	title = {First Estimates of Inequality Benchmark Incomes for a Range of Countries},
	volume = {16},
	year = {2021}}

@article{HothornHornikZeileis2006,
	author = {Hothorn, Torsten and Hornik, Kurt and Zeileis, Achim},
	doi = {10.1198/106186006X133933},
	journal = {Journal of Computational and Graphical Statistics},
	number = {3},
	pages = {651--674},
	title = {Unbiased Recursive Partitioning: A Conditional Inference Framework},
	volume = {15},
	year = {2006}}

@article{StroblEtAl2007,
	author = {Strobl, Carolin and Boulesteix, Anne-Laure and Zeileis, Achim and Hothorn, Torsten},
	doi = {10.1186/1471-2105-8-25},
	journal = {BMC Bioinformatics},
	pages = {25},
	title = {Bias in Random Forest Variable Importance Measures: Illustrations, Sources and a Solution},
	volume = {8},
	year = {2007}}

\newpage\section{Appendix A: Construction of income and consumption aggregates}\label{app:cons_inc_aggregate}

This section provides additional detail on the construction of the benchmark income and consumption aggregates used in the empirical analysis.

\subsection*{A1. Income data}
\label{app:income_data}

Household income is constructed from six main components: labour income, agricultural crop income, livestock income, non-farm enterprise income, remittances and transfers, and other income sources. All components are converted to annual values and aggregated at the household level.

\paragraph{Labour income.} Labour income is constructed using information from the labour modules of the GHS-Panel, which collect information on wage and salaried employment activities, occupation, sector of employment, payment frequency, hours worked, and remuneration received in cash and in kind. Let \(Y_h^{L}\) denote labour income of household \(h\). Labour income is computed as:
\begin{equation}
Y_{h}^{L}
=
\sum_{i \in h}
\left(
Y_{i}^{C}
+
Y_{i}^{IK}
\right),
\label{app:eq_labor_income}
\end{equation}
where \(Y_i^{C}\) denotes annual cash earnings received by individual \(i\), and \(Y_i^{IK}\) denotes the annualised value of in-kind payments received through wage employment.

The survey records labour earnings using different payment frequencies, including hourly, daily, weekly, biweekly, monthly, quarterly and annual remuneration. To ensure comparability, all earnings are converted to annual values. Hourly earnings are annualised using information on usual weekly working hours, while daily earnings are annualised assuming six working days per week. In-kind remuneration is valued using the monetary amounts reported by households. Labour supplied to household farms, household enterprises, apprenticeships and unpaid family activities is not directly monetised within this component.

\paragraph{Non-farm enterprise income.} Non-farm enterprise income is constructed from the post-harvest household questionnaire, since the non-farm enterprise module is collected only during the post-harvest visit in Wave 4. The module covers household businesses and income-generating activities outside agriculture, including petty trade, transport services, handicrafts, manufacturing, food processing, retail commerce and other informal self-employment activities.

Household non-farm enterprise income is defined as net enterprise income during the reference period. Let \(Y_h^{NF}\) denote non-farm enterprise income of household \(h\). This component is computed as:
\begin{equation}
Y_{h}^{NF}
=
\sum_{e \in h}
\left(
R_{eh} - C_{eh}
\right),
\label{app:eq_nonfarm_income}
\end{equation}
where \(R_{eh}\) denotes gross revenue generated by enterprise \(e\), and \(C_{eh}\) denotes operating costs, including expenditures on raw materials, transport, utilities, hired labour, maintenance and other business-related expenses. Enterprise income is measured at the enterprise level and reflects joint returns to labour, capital and entrepreneurial activity.

\paragraph{Agricultural income.} Agricultural income is constructed using the post-harvest agricultural modules, which collect information on crop production, expected harvests and agricultural input expenditures. Let \(Y_h^{A}\) denote agricultural income of household \(h\). Agricultural income is defined as the net value of crop production:
\begin{equation}
Y_{h}^{A}
=
V_{h}^{A}
-
C_{h}^{A},
\label{app:eq_ag_income}
\end{equation}
where \(V_h^{A}\) denotes the gross value of agricultural production and \(C_h^{A}\) denotes total agricultural input expenditures.

The value of agricultural production is constructed from the crop harvest modules. For each crop cultivated by the household, the survey records the estimated value of the quantity already harvested from each plot. Let \(V_{hpc}^{H}\) denote the reported value of harvested crop \(c\) from plot \(p\), and \(Q_{hpc}^{H}\) the corresponding quantity harvested, converted into standardised units. The crop-specific unit value is computed as:
\begin{equation}
P_{hpc}
=
\frac{V_{hpc}^{H}}
     {Q_{hpc}^{H}}.
\label{app:eq_crop_price}
\end{equation}

Where part of the crop remains unharvested at the time of interview, the expected remaining harvest is valued as:
\begin{equation}
V_{hpc}^{E}
=
P_{hpc}
\times
Q_{hpc}^{E},
\label{app:eq_expected_harvest}
\end{equation}
where \(Q_{hpc}^{E}\) denotes the expected quantity still to be harvested, also expressed in standardised units.\footnote{Crop quantities may be reported under different conditions, such as shelled versus unshelled maize or paddy versus processed rice. To ensure consistency in valuation, expected production is imputed only when the condition of the harvested and expected crop quantities coincides.}

Total crop production value is therefore defined as:
\begin{equation}
V_{h}^{A}
=
\sum_{p,c}
\left(
V_{hpc}^{H}
+
V_{hpc}^{E}
\right).
\label{app:eq_crop_value}
\end{equation}

Agricultural production costs include purchased inputs, such as fertilisers, pesticides and seeds; rental costs of agricultural land; rental costs of equipment and machinery; and transport costs associated with input acquisition.\footnote{Cash rental payments reported for a six-month period are annualised by multiplying the observed payment by two. Payments associated with rainy-season and dry-season contracts are treated as covering a full agricultural cycle and are therefore not further annualised.} Since the survey does not provide sufficient information to fully value unpaid family labour, land services or capital depreciation, this measure should be interpreted as a net cash-based approximation of agricultural income rather than full economic profit.

\paragraph{Livestock income.}

Livestock income is constructed using the livestock production modules, which collect information on animal sales, milk production, egg production and livestock-related expenditures. Let \(Y_h^{LV}\) denote livestock income of household \(h\). This variable is defined as:
\begin{equation}
Y_{h}^{LV}
=
R_{h}^{LV}
-
C_{h}^{LV},
\label{app:eq_livestock_income}
\end{equation}
where \(R_h^{LV}\) represents gross livestock revenues and \(C_h^{LV}\) denotes livestock production expenditures.

Gross livestock revenues include live animal sales, slaughtered animal sales, and revenues derived from livestock products such as milk and eggs. Where livestock products are consumed within the household, values are imputed using household-specific unit values derived from observed sales where available, or median unit values among households reporting positive sales. Egg income is measured conservatively using only the reported value of egg sales, since the survey does not identify the final destination of eggs not sold.

Livestock production costs include vaccination, veterinary services, water, purchased feed, hired labour, compensation payments and other livestock-related operating costs. Purchases of animals are treated as asset acquisitions and are not deducted as current production costs. The current value of owned livestock is also excluded to avoid conflating income flows with wealth holdings.

\paragraph{Remittances, transfers and other income.}

Remittances and private transfers include domestic and international transfers received by the household in cash or in kind. Domestic transfers are reported in Nigerian Naira. Foreign remittances reported in U.S. dollars, euros or pounds sterling are converted into Nigerian Naira using average annual exchange rates for 2018.\footnote{Foreign remittances are converted using average exchange rates of 360 NGN/USD, 424 NGN/EUR and 482 NGN/GBP, respectively.} Transfers received in kind are incorporated using the monetary values reported by respondents.

Other income sources include rental income, pensions, returns on savings and investments, social assistance programmes and miscellaneous receipts. Rental income includes both agricultural and non-agricultural property income. Agricultural rental income includes cash and in-kind payments received for renting out agricultural plots, while non-agricultural rental income includes earnings from residential or commercial property. Social assistance includes government support programmes, food assistance and other transfers received by household members, valued using reported monetary amounts where available.

\subsection*{A2. Consumption data}
\label{app:consumption_data}

Household consumption is constructed as the total monetary value of food consumption, non-food expenditure and housing services:
\begin{equation}
C_h
=
C_h^{F}
+
C_h^{NF}
+
C_h^{H},
\label{app:eq_consumption_aggregate}
\end{equation}
where \(C_h^{F}\) denotes food consumption, \(C_h^{NF}\) denotes non-food consumption expenditure, and \(C_h^{H}\) denotes housing-related consumption.

\paragraph{Food consumption.}

The food consumption modules, collected in both post-planting and post-harvest visits, include 16 food categories and more than 160 individual food items. Food consumption includes purchases, own-produced food, gifts and meals consumed outside the home. Since households report food quantities in multiple standard and local units, quantities are first converted into standardised units using the conversion factors available in the survey.

Food values are estimated using the quantity purchased in the most recent purchase and the corresponding reported value. Where household-specific purchase information is unavailable, national average prices are assigned, taking into account the size of the product where this information is available. The final food aggregate includes the value of purchased food, own-produced food and food received as gifts, valued at market prices.

Expenditure on meals consumed outside the home is reported for the previous seven days and includes breakfast, lunch, dinner, side dishes, dairy-based beverages, vegetables, and alcoholic and non-alcoholic drinks. These expenditures are annualised using a factor of \(365/7\).

\paragraph{Non-food consumption.}

Non-food consumption is constructed from a broad set of expenditure items grouped into 13 categories. The survey records these expenditures over different recall periods, including the last seven days, last 30 days, last six months and last 12 months. Expenditures are annualised using the corresponding recall-period conversion factors.

Items that do not correspond to current consumption are excluded. These include mortgages, donations, production inputs and highly irregular ceremonial expenditures. Education expenditure includes school fees, uniforms, textbooks, meals and lodging, transport, gifts to teachers, private tutoring and other school-related costs. Where detailed education components are unavailable, the aggregate education expenditure variable is used. Health expenditure includes consultations, medicines, laboratory exams, hospitalisation charges and transport to health facilities. Health expenses reported for the previous four weeks are annualised using a factor of \(52/4\).

\paragraph{Housing.}

Housing consumption is measured as the value of housing services received by the household. For tenants, actual rent paid is used as the value of housing services. For owner-occupiers and non-market tenants, rent is imputed using a log-linear hedonic model estimated among market tenants:
\begin{equation}
\log(r_i)
=
\alpha
+
\beta_1 R_i
+
\beta_2 X_i
+
\epsilon_i,
\label{app:eq_loglin}
\end{equation}
where \(r_i\) denotes reported rent for dwelling \(i\), \(R_i\) is a vector of regional indicators, \(X_i\) is a vector of dwelling characteristics, including the number of rooms, wall, floor and roof materials, and sanitation type, and \(\epsilon_i\) is the error term. Parameter estimates are presented in Table \ref{tab:ols}.

\renewcommand{\thetable}{A\arabic{table}}
\setcounter{table}{0}

\begin{table}[h]
\centering
\caption{Ordinary Least Squares Estimates}
\label{tab:ols}
\begin{tabular}{l c}
\toprule
 & Estimate (Std. error) \\
\midrule
Constant & 7.2031 (0.2319)*** \\

Number of rooms & 0.0595 (0.0262)* \\

\addlinespace
\textit{Zone (Ref.: North Central)} & \\
North East & 0.5329 (0.0263)*** \\
North West & 0.1770 (0.1200) \\
South East & 0.2972 (0.0556)*** \\
South South & 0.3032 (0.0563)*** \\
South West & 0.1660 (0.0519)** \\

\addlinespace
\textit{Type of dwelling (Ref.: Separate house)} & \\
Semi-detached house & 0.0635 (0.1636) \\
Apartment/flat & 0.6825 (0.2545)** \\
Compound house & -0.1737 (0.0629)** \\
Other & -0.2046 (0.1015)* \\

\addlinespace
\textit{Walls material (Ref.: Mud)} & \\
Unburnt bricks & 0.2995 (0.0894)*** \\
Burnt bricks & 0.3409 (0.2785) \\
Cement or concrete & 0.2384 (0.0647)*** \\
Other & 0.4675 (0.3153) \\

\addlinespace
\textit{Roof material (Ref.: Corrugated iron sheets)} & \\
Cement or concrete & 0.1920 (0.2038) \\
Asbestos sheet & 0.4578 (0.0792)*** \\
Long/short span sheets & 0.2471 (0.1297). \\
Other & 0.2025 (0.3545) \\

\addlinespace
\textit{Floor material (Ref.: Sand, dirt, straw)} & \\
Smoothed mud & -0.0417 (0.2834) \\
Smooth cement/concrete & 0.5936 (0.2642)* \\
Tile & 1.1249 (0.3485)** \\
Terrazo & 0.9163 (0.4877). \\
Other & 1.4570 (0.3763)*** \\

\addlinespace
\textit{Sanitary type (Ref.: Flush to piped sewage system)} & \\
Flush to septic tank & -0.1443 (0.1488) \\
Flush to pit latrine & -0.4586 (0.1920)* \\
Pit latrine with slab & -0.7509 (0.0908)*** \\
Pit latrine without slab/open pit & -0.5269 (0.1857)** \\
Hanging toilet/hanging latrine & -0.6757 (0.0830)*** \\
No facilities, bush or field & -1.1707 (0.2352)*** \\
Other & -0.3087 (0.0511)*** \\

\midrule
Adjusted R-squared & 0.5721 \\
Number of observations & 1006 \\
F-statistic & 44.2*** \\
\bottomrule
\end{tabular}

\begin{tablenotes}
\footnotesize
\item Note: Standard errors clustered at the zone level in parentheses. *** $p < 0.001$; ** $p < 0.01$; * $p < 0.05$; . $p < 0.1$.
\end{tablenotes}
\end{table}

Because the model is estimated in logarithms, predicted values are retransformed using Duan's smearing estimator:
\begin{equation}
\hat{r}_i
=
\exp
\left(
\hat{\alpha}
+
\hat{\beta}_1 R_i
+
\hat{\beta}_2 X_i
\right)
\cdot
\widehat{E}
\left[
\exp(\epsilon_i)
\right],
\label{app:eq_Duan}
\end{equation}
where the second term is the sample average of the exponentiated residuals from the log-linear model.

\subsection*{A3. Adjustments to household income and consumption}
\label{app:adjustments}

All income and consumption components are harmonised to annual values. Nominal values are then expressed in real terms using Consumer Price Index data from the Nigerian Bureau of Statistics. These adjustments account for both temporal price differences across interview months and spatial price differences between rural and urban areas, ensuring that welfare aggregates are comparable across households interviewed at different times and in different locations.

\newpage
\section*{Appendix B}
\renewcommand{\thetable}{B\arabic{table}}
\setcounter{table}{0}
\begin{longtable}{p{6cm} c p{6cm} c}
\caption{Items Included in Food Categories}
\label{tab:food_items} \\
\toprule
\textbf{Item} & \textbf{Code} & \textbf{Item} & \textbf{Code} \\
\midrule
\endfirsthead

\caption[]{Items Included in Food Categories (continued)} \\
\toprule
\textbf{Item} & \textbf{Code} & \textbf{Item} & \textbf{Code} \\
\midrule
\endhead

\multicolumn{4}{r}{\textit{Continued on next page}} \\
\endfoot

\bottomrule
\endlastfoot

\multicolumn{4}{l}{\textbf{Grains (1)}} \\
Guinea corn/sorghum & 10 & Maize (Unshelled) & 20 \\
Millet & 11 & Maize (Shelled) & 22 \\
Rice - local & 13 & Other grains and flour & 23 \\
Rice - Imported & 14 & Yam flour & 17 \\
Maize flour & 16 & Cassava flour & 18 \\
Wheat flour & 19 & & \\

\addlinespace
\multicolumn{4}{l}{\textbf{Processed Foods \& Snacks (2)}} \\
Bread & 25 & Biscuits & 28 \\
Cake & 26 & Meat Pie/Sausage Roll & 29 \\
Buns/Pofpof/Donuts & 27 & & \\

\addlinespace
\multicolumn{4}{l}{\textbf{Roots and Tubers (3)}} \\
Cassava - roots & 30 & Sweet potatoes & 36 \\
Yam - roots & 31 & Potatoes & 37 \\
Gari - white & 32 & Other roots and tuber & 38 \\
Gari - yellow & 33 & Plantains & 35 \\
Cocoyam & 34 & & \\

\addlinespace
\multicolumn{4}{l}{\textbf{Nuts and Pulses (4)}} \\
Soya beans & 40 & Groundnuts (Shelled) & 44 \\
Brown beans & 41 & Other nuts/seeds/pulses & 45 \\
White beans & 42 & Coconut & 46 \\
Groundnuts (Unshelled) & 43 & Cashew nut & 48 \\
Kola nut & 47 & & \\

\addlinespace
\multicolumn{4}{l}{\textbf{Oils and Fats (5)}} \\
Palm oil & 50 & Other oil and Fat & 53 \\
Butter/ Margarine & 51 & Animal fat & 56 \\
Groundnuts Oil & 52 & & \\

\addlinespace
\multicolumn{4}{l}{\textbf{Fruits (6)}} \\
Bananas & 60 & Mangoes & 62 \\
Orange/tangerine & 61 & Avocado pear & 63 \\
Pineapples & 64 & Other fruits & 66 \\
Pawpaw & 67 & Watermelon & 68 \\
Apples & 69 & & \\

\addlinespace
\multicolumn{4}{l}{\textbf{Vegetables (7)}} \\
Tomatoes & 70 & Garden eggs/egg plant & 73 \\
Tomato puree (canned) & 71 & Okra - fresh/dried & 74/75 \\
Onions & 72 & Fresh/Dry Pepper & 76/77 \\
Leaves (Cocoyam, Spinach) & 78 & Other vegetables & 79 \\

\addlinespace
\multicolumn{4}{l}{\textbf{Poultry and Eggs (8)}} \\
Chicken & 80 & Agricultural eggs & 83 \\
Duck & 81 & Local eggs & 84 \\
Other domestic poultry & 82 & Other eggs (not chicken) & 85 \\

\addlinespace
\multicolumn{4}{l}{\textbf{Meat (9)}} \\
Beef & 90 & Goat & 93 \\
Mutton & 91 & Wild game/bush meat & 94 \\
Pork & 92 & Canned beef/corned beef & 95 \\
Other meat & 96 & & \\

\addlinespace
\multicolumn{4}{l}{\textbf{Fish and Seafood (10)}} \\
Fish - fresh & 100 & Fish - dried & 103 \\
Fish - frozen & 101 & Snails & 104 \\
Fish - smoked & 102 & Seafood (lobster, crab, etc) & 105 \\
Canned fish/seafood & 106 & Other fish or seafood & 107 \\

\addlinespace
\multicolumn{4}{l}{\textbf{Milk and Dairy (11)}} \\
Fresh milk & 110 & Baby milk powder & 112 \\
Milk powder & 111 & Milk tinned (unsweetened) & 113 \\
Cheese (wara) & 114 & Other milk products & 115 \\

\addlinespace
\multicolumn{4}{l}{\textbf{Coffee, Tea and Breakfast (12)}} \\
Coffee & 120 & Tea & 122 \\
Chocolate drinks (Milo) & 121 & Sugar & 130 \\
Honey & 132 & Other sweets/confectionary & 133 \\

\addlinespace
\multicolumn{4}{l}{\textbf{Miscellaneous and Spices (13)}} \\
Salt & 141 & Melon (shelled/unshelled) & 145/146 \\
Unground/Ground Ogbono & 142/143 & Other Spices & 148 \\
Ground Pepper & 144 & & \\

\addlinespace
\multicolumn{4}{l}{\textbf{Non-Alcoholic Beverages (14)}} \\
Bottled water & 150 & Soft drinks (Coca Cola, etc) & 153 \\
Sachet water & 151 & Fruit juice canned/Pack & 154 \\
Malt drinks & 152 & Other non-alcoholic drinks & 155 \\

\addlinespace
\multicolumn{4}{l}{\textbf{Alcoholic Beverages (15)}} \\
Beer (local and imported) & 160 & Pito & 162 \\
Palm wine & 161 & Gin & 163 \\
Other alcoholic beverages & 164 & & \\

\end{longtable}

\begin{longtable}{p{6cm} c p{6cm} c}
\caption{Items Included in Non-Food Categories}
\label{tab:nonfood} \\
\toprule
\textbf{Item} & \textbf{Code} & \textbf{Item} & \textbf{Code} \\
\midrule
\endfirsthead

\caption[]{Items Included in Non-Food Categories (continued)} \\
\toprule
\textbf{Item} & \textbf{Code} & \textbf{Item} & \textbf{Code} \\
\midrule
\endhead

\multicolumn{4}{r}{\textit{Continued on next page}} \\
\endfoot

\bottomrule
\endlastfoot

\multicolumn{4}{l}{\textit{Smoking}} \\
Cigarettes or tobacco & 101 & Wages paid to staff/maid/lawnsboy & 325 \\
Matches & 102 & Repairs \& maintenance to dwelling & 327 \\
 &  & Repairs to household and personal items & 328 \\
 &  & Building items -- cement, bricks, etc. & 508 \\

\addlinespace
\multicolumn{4}{l}{\textit{Utilities and Fuels}} \\
Electricity & 305 & Infant clothing & 401 \\
Light bulbs/globes & 311 & Baby nappies/diapers & 402 \\
Water & 312 & Boys tailored clothes & 403 \\
Kerosene & 301 & Boys dress (ready made) & 404 \\
Palm kernel oil & 302 & Girls tailored clothes & 405 \\
Gas (lighting/cooking) & 303 & Girls dress (ready made) & 406 \\
Other liquid cooking fuel & 304 & Men tailored clothes & 407 \\
Candle & 306 & Men dress (ready made) & 408 \\
Firewood & 307 & Women tailored clothes & 409 \\
Charcoal & 308 & Women dress (ready made) & 410 \\
Petrol & 309 & Ankara/George materials & 411 \\
Diesel & 310 & Other clothing materials & 412 \\
Lubricants (oil, grease, etc.) & 330 & Tailoring charges & 417 \\
 &  & Laundry and dry cleaning & 418 \\
 &  & Hand loomed: ASO-OKE & 431 \\

\addlinespace
\multicolumn{4}{l}{\textit{Personal Care}} \\
Soap and washing powder & 313 & Boy's shoes & 413 \\
Toilet paper & 314 & Men's shoes & 414 \\
Personal care goods & 315 & Girl's shoes & 415 \\
Vitamin supplements & 316 & Lady's shoes & 416 \\
Insecticides, disinfectant, cleaners & 317 & Repairs of footwear & 432 \\

\addlinespace
\multicolumn{4}{l}{\textit{Recreation and Communication}} \\
Books (not for school) & 426 & Bowls, glassware, plates, silverware & 419 \\
Night lodging (house/hotel) & 428 & Cooking utensils & 420 \\
Newspapers and magazines & 103 & Cleaning utensils (brooms, brushes) & 421 \\
Gambling, lotto, raffles & 105 & Torch/flashlight & 422 \\
Cinema, DVD rental & 322 & Umbrella and raincoat & 423 \\
Sports equipment, musical instruments, toys & 506 & Paraffin lamp & 424 \\
Film, processing, camera & 507 & Stationery (not for school) & 425 \\
Postal (stamps, courier) & 318 & House decorations & 427 \\
Recharge cards & 319 & Electric kettle & 433 \\
Landline charges & 320 & Coal pot/non-electric appliance & 434 \\
Internet services & 321 & Repairs of appliances & 435 \\
Cell phone handset & 440 & Bed sheets, blankets & 436 \\
Personal computer & 441 & Pillow & 437 \\
 &  & Curtains and other linen & 438 \\
 &  & Carpet and floor covering & 439 \\
 &  & Carpet, rugs, drapes & 501 \\
 &  & Linen (towels, sheets, blankets) & 502 \\
 &  & Sleeping mat & 503 \\
 &  & Mosquito net & 504 \\
 &  & Mattress & 505 \\

\addlinespace
\multicolumn{4}{l}{\textit{Transport and Insurance}} \\
Public transport (commuting) & 104 & Health insurance & 510 \\
Motor vehicle service/repair/parts & 323 & Auto insurance & 511 \\
Bicycle service/repair/parts & 324 & Home insurance & 512 \\
 &  & Life insurance & 513 \\

\addlinespace
\multicolumn{4}{l}{\textit{Taxes and Fines}} \\
Council rates & 509 & Fines or legal fees & 514 \\

\end{longtable}

\end{document}